\pdfoutput=1

\documentclass[11pt]{article}
\usepackage[table]{xcolor}
\usepackage{ACL2023}

\usepackage{times}
\usepackage{latexsym}

\usepackage[T1]{fontenc}

\usepackage[utf8]{inputenc}

\usepackage{microtype}

\usepackage{inconsolata}
\usepackage{xcolor}
\usepackage{tcolorbox}
\usepackage{booktabs}
\usepackage{textgreek}
\usepackage{mathtools}
\usepackage{amsmath}
\usepackage{txfonts}
\usepackage{multirow}
\usepackage{enumitem}
\usepackage{xspace}
\usepackage{pgfplots}
\usepackage{pgfplotstable}
\usepackage{tikz}
\newcommand{\macrof}{$\mathrm{m}$$\mathrm{F_1}$\xspace}
\newcommand{\microf}{$\mathrm{\muup}$$\mathrm{F_1}$\xspace}
\newenvironment{myquote}[1]%
  {\list{}{\leftmargin=#1\rightmargin=#1}\item[]}%
  {\endlist}

\title{LeXFiles and LegalLAMA: Facilitating English Multinational \\ Legal Language Model Development}

\author{Ilias Chalkidis\thanks{\hspace{0.5em}Equal contribution.}\qquad Nicolas Garneau$^\ast$ \qquad Anders Søgaard \\
Department of Computer Science, University of Copenhagen, Denmark \\ \AND
Cătălina Goanță \\ 
  Utrecht University School of Law, \\ Netherlands \\ \And
  Daniel Martin Katz \\
  Illinois Tech – Chicago Kent College of Law, \\ IL, United States
  }

\begin{document}
\maketitle

\begin{abstract}
In this work, we conduct a detailed analysis on the performance of legal-oriented pre-trained language models (PLMs).
We examine the interplay between their original objective, acquired knowledge, and legal language understanding capacities which we define as the upstream, probing, and downstream performance, respectively.
We consider not only the models' size but also the pre-training corpora used as important dimensions in our study.
To this end, we release a multinational English legal corpus (\textsc{LeXFiles}) and a legal knowledge probing benchmark (\textsc{LegalLAMA}) to facilitate training and detailed analysis of legal-oriented PLMs. We release two new legal PLMs trained on \textsc{LeXFiles} and evaluate them alongside others on \textsc{LegalLAMA} and \textsc{LexGLUE}.
We find that probing performance strongly correlates with upstream performance in related legal topics.
On the other hand, downstream performance is mainly driven by the model's size and prior legal knowledge which can be estimated by upstream and probing performance. Based on these findings, we can conclude that both dimensions are important for those seeking the development of domain-specific PLMs.
\end{abstract}

\section{Introduction}

Following closely the advances in the development of NLP technologies, the legal NLP literature is flourishing with the release of many new resources, including large legal corpora \cite{hendersonkrass2022pileoflaw}, datasets \cite{chalkidis-etal-2021-multieurlex, koreeda-manning-2021-contractnli-dataset, zhengguha2021,chalkidis-etal-2022-lexglue,habernal-etal-2022-argument}, and pre-trained legal-oriented language models (PLMs) \cite{chalkidis-etal-2020-legalbert, zhengguha2021,xiao-etal-2021}. Benchmark suites \cite{chalkidis-etal-2022-lexglue,hwang2022a,niklaus2023lextreme} to evaluate the performance of PLMs in a more systematic way have been also developed, showcasing the superiority of legal-oriented PLMs over generic ones on downstream legal NLP tasks.

\begin{table*}
\centering
    \resizebox{0.95\textwidth}{!}{
    \begin{tabular}{rr|r|c}
\bf Sub-Corpus (Source)  & \bf \# Documents & \bf \# Tokens / Percentage (\%)   & \bf Sampling Smoothing (\%)  \\
\cmidrule(r){2-4}
\rowcolor{gray!20} EU Legislation               &  93.7K     & 233.7M (01.2\%)   & 05.0\%          \\    
EU Case Law                  &  29.8K     & 178.5M (00.9\%)   & 04.3\%       \\   
\rowcolor{gray!20} UK Legislation               &  52.5K     & 143.6M (00.7\%)   & 03.9\%   \\    
UK Case Law                  &  47K       & 368.4M (01.9\%)   & 06.2\%         \\   
\rowcolor{gray!20} Canadian Legislation         &  6K        & 33.5M (00.2\%)  & 01.9\%     \\ 
Canadian Case Law            &  11.3K     & 33.1M (00.2\%)   & 01.8\%         \\ 
\rowcolor{gray!20} U.S. Legislation             &  518       & 1.4B (07.4\%)   & 12.3\%          \\ 
U.S. Case Law                &  4.6M      & 11.4B (59.2\%)  & 34.7\%       \\ 
\rowcolor{gray!20} U.S. Contracts               &  622K      & 5.3B (27.3\%)  & 23.6\%          \\ 
 ECtHR Case Law               &  12.5K     & 78.5M (00.4\%)   & 02.9\%     \\    
\rowcolor{gray!20} Indian Case Law              &  34.8K     & 111.6M (00.6\%)   & 03.4\%      \\  
\cmidrule(r){2-4}
\bf Total                        & 5.8M      & 18.8B (100\%)   & 100\%           \\ 
\end{tabular}
}
\caption{Core statistics of the newly introduced \textsc{LeXFiles} corpus. In the last column, we present the sampling smoothing percentages used to train our LexLM models (Section~\ref{sec:models}).}
\label{tab:pile}
\vspace{-2mm}
\end{table*}

Despite this impressive progress, there is still not a thorough study on (a) how PLMs trained under different settings (pre-training corpora, size of the model) perform across different legal sub-corpora, and (b) what sort of knowledge such models have acquired from pre-training, and (c) how important is domain (legal) specificity vs general (cross-domain) legal knowledge.
Furthermore, often times, legal NLP relies on datasets without drawing clear lines and comparisons between the various legal systems they may reflect. A legal system may be defined as a set of rules adopted and enforced at a given governance level, which may be national, regional or international~\cite{Friedman2017}, e.g., UK, EU, US, CoE, etc. 

We define the upstream evaluation as the task PLMs are explicitly designed to do: Masked Language Modelling (MLM)~\citep{devlin-etal-2019-bert}.
We then probe for specific legal concepts that are legal-system specific, in a similar fashion as \citet{petroni-etal-2019-language} did using the ``LAnguage Models Analysis'' (LAMA) framework.
Finally, we assess the PLMs performance in LexGLUE~\cite{chalkidis-etal-2022-lexglue} downstream tasks.
More importantly, we explore how the aforementioned factors (upstream, and probing performance) interplay and relate to downstream performance. Our contributions are:
\vspace{-2mm}
\begin{enumerate}[label=(\alph*),leftmargin=14pt,itemsep=2pt,parsep=2pt] 
    \item We release \textsc{LeXFiles}, a new diverse English legal corpus including 11 sub-corpora that cover legislation and case law from 6 primarily English-speaking legal systems (EU, CoE, Canada, US, UK, India). The corpus comprises approx.~6 million documents which sum up to approx.~19 billion tokens. 
    \item We release 2 new legal-oriented PLMs, dubbed LexLMs, warm-started from the RoBERTa~\cite{liu-2019-roberta} models, and further pre-trained on the \textsc{LeXFiles} for 1M additional steps.
    \item We release \textsc{LegalLAMA}, a diverse probing benchmark suite comprising 8 sub-tasks that aims to assess the acquaintance of legal knowledge that PLMs acquired in pre-training. 
    \item We evaluate 7 PLMs 
    on both \textsc{LeXFiles} and \textsc{LegalLAMA}, analyzing their performance out of the box per \textsc{LeXFiles} sub-corpus and \textsc{LegalLAMA} tasks. We also fine-tune and evaluate these models in selected \textsc{LexGLUE} tasks, and examine the interplay between MLM, probing, and downstream performance.
\end{enumerate}

\section{LeXFiles Corpus}
\label{sec:corpus}

The \textsc{LeXFiles} is a new diverse English multinational legal corpus that we created including 11 distinct sub-corpora (Table~\ref{tab:pile}) that cover legislation and case law from 6 primarily English-speaking legal systems (EU, CoE, Canada, US, UK, India).
The corpus contains approx.~19 billion tokens.
In comparison, the \textsc{Pile of Law} corpus released by \citet{hendersonkrass2022pileoflaw} comprises 32 billion in total, where the majority (26/30) of sub-corpora come from the United States of America (USA), hence the corpus as a whole is biased towards the US legal system in general, and the federal or state jurisdiction in particular, to a significant extent.
The \textsc{LeXFiles}'s sub-corpora are:\vspace{-2mm}

\begin{enumerate}[label=(\alph*),itemsep=-0.2em]
\item \emph{EU Legislation.} We release 93.7K EU laws (regulations, decisions, directives) published in EUR-Lex, the website of the EU Publication Office.\footnote{\url{https://eur-lex.europa.eu/}}

\item \emph{EU Case Law.} We release 29.8K EU court decisions, mainly issued from the Court of Justice (CJEU), published in EUR-Lex.\footnotemark[1]

\item \emph{UK Legislation.} We release 52.5 UK laws published in UK.LEGISLATION.GOV.UK, the official website of the UK National Archives.\footnote{\url{https://www.legislation.gov.uk/}}

\item \emph{UK Case Law.} We release 47K UK court decisions published in the British and Irish Legal Information Institute (BAILII) database.\footnote{\url{https://www.bailii.org/}}

\item \emph{US Legislation.} We re-distribute 518 US state statutes (legislation) originally published by \citet{hendersonkrass2022pileoflaw}.

\item \emph{US Case Law.} We release 4.6M US decisions (opinions) published by Court Listener,\footnote{\url{https://www.courtlistener.com/}} a web database hosted by the Free Law Project.\footnote{We release decisions published from 1965 on-wards (cf. post Civil Rights Act), as a hard threshold for cases that possibly rely on out-dated and discriminatory law standards. The rest of the sub-corpora include more recent documents.}

\item \emph{US Contracts.} We release 622K US contracts (agreements) obtained from
US Securities and Exchange Commission (SEC) filings, which are publicly available from the SEC-EDGAR\footnote{\url{https://www.sec.gov/edgar}} database.

\item \emph{Canadian Legislation.} We release 6K Canadian laws (acts, regulations) published in the official legislation portal of Canada.\footnote{\url{https://laws-lois.justice.gc.ca/eng/}}

\item \emph{Canadian Case Law.}  We re-distribute 13.5K Canadian decisions (opinions) originally published by \citet{hendersonkrass2022pileoflaw}.

\item \emph{ECtHR Case Law.} We release 12.5K decisions ruled by the European Court of Human rights (ECtHR) published in HUDOC,\footnote{\url{https://hudoc.echr.coe.int/eng}} the database of ECtHR.

\item \emph{Indian Case Law.} We include 34.8K Indian Supreme Court cases originally published by \citet{malik-etal-2021-ildc}.

\end{enumerate}

The \textsc{LeXFiles} is pre-split into training and test subsets to provide a fair ground for comparing the performance of PLMs that have not been trained in the training set. We use the training subset of the \textsc{LeXFiles} corpus to train 2 new transformer-based languages models, dubbed \textsc{LexLMs} (Section~\ref{sec:models}), and evaluate their MLM performance across many other already available PLMs (Section~\ref{sec:mlm}).

\section{\textsc{LegalLAMA} Benchmark}
\label{sec:legallama}

LAnguage Model Analysis (LAMA)~\citep{petroni-etal-2019-language} is a probing task that is designed to assess specific capabilities of PLMs.
The general framework of LAMA is to let PLMs predict a target token behind a \texttt{[MASK]} given its context, e.g., \emph{``Paris is the capital of} \texttt{[MASK]}\emph{''}, where the answer is `France'.
\textsc{LegalLAMA} is a new probing benchmark suite inspired by this framework.
It includes 8 sub-tasks that aim to assess the acquaintance of legal knowledge that PLMs acquired in the pre-training phase in a \emph{zero-shot fashion}. Such tasks cannot be resolved by laypersons or even law professionals that are not experts in the specific fields of law in many cases.\footnote{In Appendix~\ref{sec:legalama_discuss}, we present a discussion on the \textsc{LegalLAMA} tasks' level of difficulty.} The acquaintance of legal knowledge can be interpreted as some form of primitive understanding of the law, for specific aspects in very controlled (limited) settings -limited legal concepts under a specific jurisdiction-. As \citet{sahlgren-2021} mentioned:
\begin{myquote}{0.1in}
``Rather than asking whether a language model understands or not, we should ask \emph{to what extent, and in which way}, a model understands.''
\end{myquote}

We further extend the LAMA framework by allowing PLMs to predict multi-token targets.
Take for example the ``\textit{Drug Trafficking}'' offence under the ``\textit{Drug-Related}'' crimes of the US legislation.
Using the RoBERTa tokenizer, this term is split into two tokens, that is ``\textit{Drug}'' and ``\textit{Trafficking}''.
We replace thus the ``drug trafficking'' phrase with two \texttt{[MASK]} tokens, and then ask the model to predict these tokens simultaneously.
\begin{figure}[h!]
    \centering
    \includegraphics[width=0.9\columnwidth]{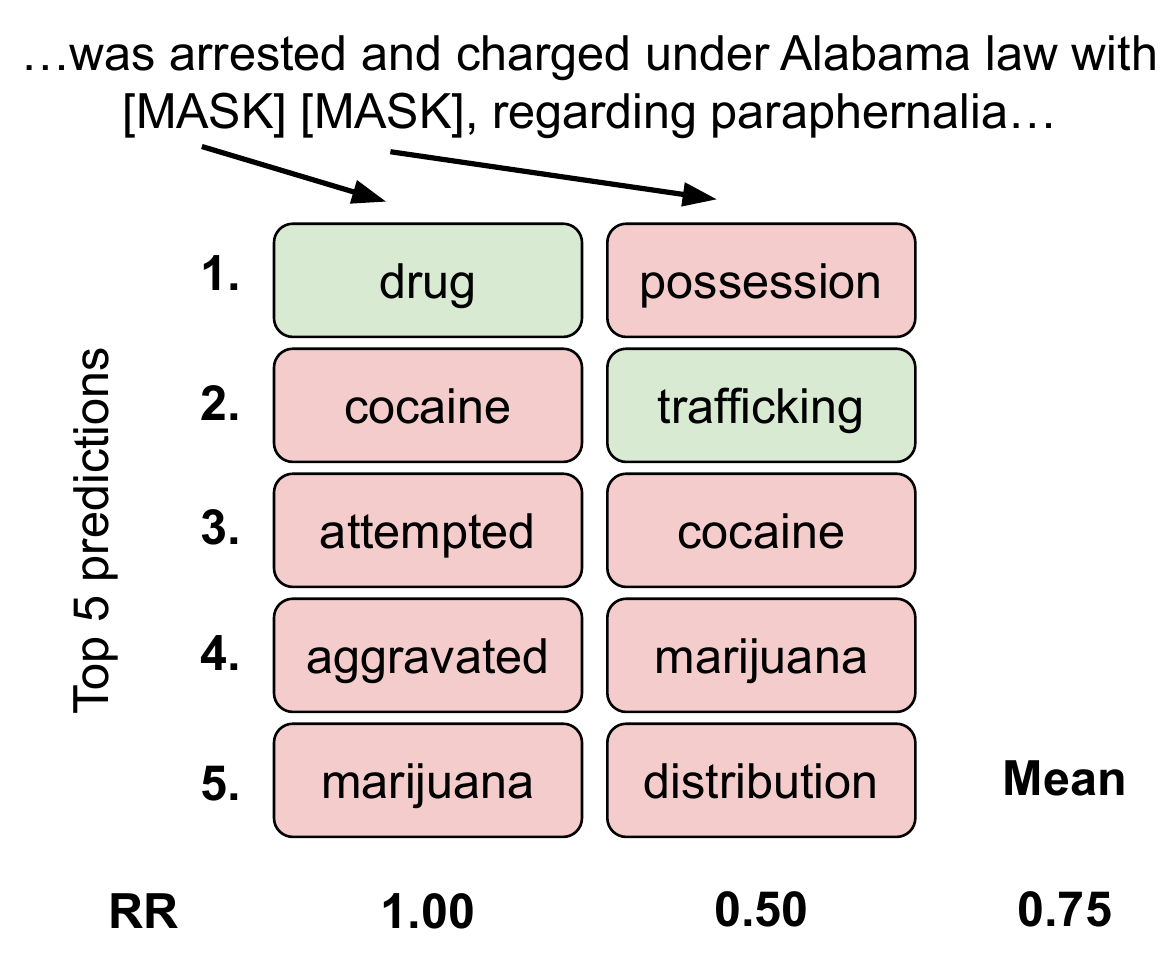}
    \vspace{-2mm}
    \caption{Example from the `Terminology (US)' sub-task. Multi-token LAMA where ``drug trafficking'' has been replaced with two \texttt{[MASK]} tokens. Given the rankings of each predicted token, we compute the reciprocal rank (RR) and obtain a mean reciprocal rank (MRR) over the \texttt{[MASK]} tokens.}
    \label{fig:multitoken_lama}
    \vspace{-4mm}
\end{figure}

We evaluate the overall performance of PLMs using the macro-averaged Mean Reciprocal Rank (MRR)~\citep{207141} over the set of labels (not the entire vocabulary).\footnote{We decided to report only MRR results in the main paper for the sake of clarity. Moreover, MRR avoids penalizing for near-identical outcomes. Detailed results including Precision at 1 (P@1) are available in Appendix~\ref{app:detailed_lama}.}
In the case of multi-token targets, we average the MRR over the predicted tokens.\footnote{A stricter evaluation would be to consider a multi-token prediction valid only if all the sub-tokens are properly predicted by the PLM. We decided to average the MRR to consider minor variations and errors.}
Note that \textsc{LegalLAMA} examples come from the test subset of the related LexFiles sub-corpora in order to have a fair comparison between models trained or not on the LexFiles training sets.
We provide a concrete example in Figure~\ref{fig:multitoken_lama}, and describe the tasks in detail:\vspace{-1mm}

\paragraph{ECHR Articles (CoE).} 
In this task, we have paragraphs from the court assessment section of ECtHR decisions. We extract those paragraphs from the newly introduced ECHR corpus presented in Section~\ref{sec:corpus}. The paragraphs include references to ECHR articles, e.g., \emph{``Article \texttt{[MASK]} of the Convention''}, where \texttt{[MASK]} is the article number. For example, \emph{``The applicant complained under Article \textbf{[2]} of the Convention that the prison authorities had failed to protect her son's right to life by taking the necessary measures.''} Given a paragraph, where the article number is masked, the model has to predict the associated article number given the context. The dataset is composed of 5,072 test instances containing on average 69 tokens and 13 unique article numbers to predict.

\paragraph{Contractual Section Titles (US).} In this task, we have sections from US contracts reusing the dataset of \citet{tuggener-etal-2020-ledgar}. Contractual sections are usually numbered and titled, e.g., \emph{"10. \textbf{[Arbitration]}. Any controversy, dispute or claim directly or indirectly arising out of or relating to this Agreement [...]"}. The section titles reflect the content (subject matter) of the section, and are commonly re-used. Given a section, where the section title is masked, the model has to predict the associated title given the context. The dataset is composed of 1,527 test instances containing on average 85 tokens and 20 unique section titles to predict.

\paragraph{Contract Types (US).}  In this task, we have introductory paragraphs from US contracts. We extract those paragraphs from the newly introduced corpus of US contracts, presented in Section~\ref{sec:corpus}. Introductory paragraphs usually start with the contract title revealing the contract type, e.g., \emph{"Service Agreement"}, and follow with the names of the involved parties, and their roles in this agreement. For example, \emph{``This \textbf{[Purchase]} Agreement is entered into this 23rd day of January 2020 by and between A (the "Purchaser") and B (the "Seller").''}.
Given an introductory paragraph, where the contract type is masked, the model has to predict the associated type given the context. The task is composed of 1,089 test instances containing on average 150 tokens and 15 unique types of contracts to predict.

\paragraph{Crime Charges (US).}  In this task, we have paragraphs from US court judgments (opinions). We extract those paragraphs from the US case law corpus, presented in Section~\ref{sec:corpus}. We select a list of criminal offenses (e.g., ``Sexual Assault''), categorized into 11 major categories (e.g., Sex-related) from the FindLaw website.\footnote{\url{https://www.findlaw.com/criminal/criminal-charges.html}} We filter out paragraphs that refer the specified criminal charges verbatim. For example, \emph{``A person commits the crime of \textbf{[burglary]} in the first degree when he or she enters or remains unlawfully in a building with the intent to commit a crime against a person or property therein''}
Given a paragraph, where a criminal charge is masked, the model has to predict the associated criminal charge given the context. The task is composed of 4,518 test instances containing on average 118 tokens and 59 charges to predict.

\paragraph{Legal Terminology (US).}  In this task, we have paragraphs from US court judgments (opinions). We extract those paragraphs from the US case law corpus, presented in Section~\ref{sec:corpus}. We select a subset of legal terms per legal topic (e.g., finance law, property law, family law) using the legal vocabularies provided by the Legal Information Institute (LII) of the Cornell Law School.\footnote{\url{https://www.law.cornell.edu/}} We filter out paragraphs that use the specified legal terms. For example, \emph{``The \textbf{[marital privilege]} against self-incrimination is [...] grounded upon the theory that just as one may not be convicted by his own compelled testimony, so may he not be convicted by the testimony of his spouse.''}
Given a paragraph, where a legal term is masked, the model has to predict the associated legal term given the context. The task is composed of 5,829 test instances containing on average 308 tokens and 92 legal terms from 7 topics to predict.

\paragraph{Legal Terminology (EU).}  In this task, we have paragraphs from CJEU judgments (opinions). We extract those paragraphs from the newly introduced EU case law corpus, presented in Section~\ref{sec:corpus}. We select a subset of legal terms based on the subject matters provided by the database of the courts (CURIA).\footnote{\url{https://curia.europa.eu/}} We filter out paragraphs that use the specified legal terms. For example, \emph{``The guiding principle at the basis of EU \textbf{[data protection]} law is that of a self-determined decision of an individual who is capable of making choices about the use and processing of his or her data.''}
Given a paragraph, where a legal term is masked, the model has to predict the associated legal term given the context. The task is composed of 2,127 test instances containing on average 164 tokens and 42 legal terms from 23 topics to predict.

\begin{table*}[t]
    \centering
    \resizebox{\textwidth}{!}{
    \begin{tabular}{ll|c|c|c|ll}
         \multicolumn{2}{l|}{\bf Model (Source)} & \bf \# Params & \bf \# Vocab & \bf \# Acc. Tokens & \multicolumn{2}{c}{\bf Pre-training Corpora} \\
         \midrule
         RoBERTa & \cite{liu-2019-roberta} & 124/355M & 50K & 2T & (160GB) & Generic Corpora  \\
         LegalBERT & \cite{chalkidis-etal-2020-legalbert} & 110M & 32K & 43B & (12GB) & Legal Corpora \\
         CaseLawBERT & \cite{zhengguha2021} & 110M & 32K & 43B & (37GB) & US Case Law \\
         PoL-BERT & \cite{hendersonkrass2022pileoflaw} & 340M & 32K & 130B & (256GB) & US Legal Corpora \\
         LexLM & (ours) & 124/355M & 50K & 2T + 256B & (175GB) & Legal Corpora \\
    \end{tabular}
    }
    \vspace{-2mm}
    \caption{Key specifications of the examined models. We report the number of parameters, the size of vocabulary, the number of accumulated training tokens, and the nature of pre-trainig corpora.}
    \label{tab:examined_models}
\end{table*}

\paragraph{Legal Terminology (CoE).}  In this task, we have paragraphs from ECtHR decisions. We extract those paragraphs from the newly introduced ECHR corpus presented in Section~\ref{sec:corpus}. We select a subset of legal terms (legal issues) based on the keywords provided by the database of the courts (HUDOC).\footnote{\url{https://www.echr.coe.int/Documents/HUDOC_Keywords_ENG.pdf}} We filter out paragraphs that use the specified legal terms. For example, \emph{``The applicants alleged that their relatives'  \textbf{[right to life]} was violated in that they were deliberately killed by village guards.''}
Given a paragraph, where a legal term is masked, the model has to predict the associated legal term given the context. The task is composed of 6,803 test instances containing on average 97 tokens and 250 legal terms from 15 articles to predict.

\paragraph{Criminal Code Sections (Canada).}  In this task, we have paragraphs from the Criminal Court of Canada's decisions containing Section Numbers of the Criminal Code of Canada (CCC)\footnote{\url{https://laws-lois.justice.gc.ca/eng/acts/c-46/index.html}}. For example, \emph{``Section \textbf{[680]} of the Criminal Code provides that a bail review is to be conducted by a panel of this court where directed by the Chief Justice.''}
Given a paragraph, where a criminal code's section is masked, the model has to predict the associated section number, paragraph, and sub-paragraph (if any)  given the context. The task is composed of 321 test instances containing on average 72 tokens and 144 different section numbers to predict.\vspace{1mm}

\noindent In Appendix~\ref{sec:legalama_lists}, we present the full list of vocabulary (masked terms) grouped in categories (clusters) -when applicable- per \textsc{LegalLAMA} sub-task.

\begin{table*}[t]
    \centering
    \resizebox{\textwidth}{!}{
    \begin{tabular}{rc|c|c|c|c|c|c}
               \bf Sub-Corpus &                     RoBERTa-B &       RoBERTa-L &             LegalBERT &                   CL-BERT &                                           PoL-BERT &           LexLM-B &          LexLM-L \\
           \cmidrule(r){2-8}
           EU Legislation & \cellcolor{gray!20}72.0  & 75.1  & \cellcolor{gray!20} \bf 83.1 & 61.4 & \cellcolor{gray!20} 73.3 & 78.7 & \cellcolor{gray!20} 81.8 \\
           EU Case Law & \cellcolor{gray!20}72.7 & 76.5& \cellcolor{gray!20} 81.4 & 63.0 & \cellcolor{gray!20} 68.5 & 79.8 & \cellcolor{gray!20} \bf 82.9 \\
           \cmidrule(r){2-8}
           UK Legislation & \cellcolor{gray!20}71.3 & 75.1 & \cellcolor{gray!20} 86.2 & 65.1 & \cellcolor{gray!20} 72.8 & 84.1 & \cellcolor{gray!20} \bf 87.3 \\
           UK Case Law & \cellcolor{gray!20}68.9 & 73.2 & \cellcolor{gray!20} 72.3 & 61.2 & \cellcolor{gray!20} 62.4 & 73.2 & \cellcolor{gray!20} \bf 76.9 \\
           \cmidrule(r){2-8}
     CAN Legislation & \cellcolor{gray!20}75.5 & 78.9 & \cellcolor{gray!20} 80.6 & 66.4 & \cellcolor{gray!20} 73.3 & 82.9 & \cellcolor{gray!20} \bf 85.2 \\
     CAN Case Law & \cellcolor{gray!20}62.8 & 66.0 & \cellcolor{gray!20} 73.8 & 64.1 & \cellcolor{gray!20} 66.0 & 76.7 & \cellcolor{gray!20} \bf 80.3 \\
     \cmidrule(r){2-8}
           US Case Law & \cellcolor{gray!20}68.2 & 72.5  & \cellcolor{gray!20} 71.6 & 64.4 & \cellcolor{gray!20} 63.8 & 71.7 & \cellcolor{gray!20} \bf  74.8 \\
           US Legislation & \cellcolor{gray!20}74.5 & 78.1  & \cellcolor{gray!20} 79.7 & 65.3 & \cellcolor{gray!20} 77.0 & 80.5 & \cellcolor{gray!20} \bf 83.5 \\
            US Contracts & \cellcolor{gray!20}67.5 & 70.9  & \cellcolor{gray!20} \bf 89.1 & 69.5 & \cellcolor{gray!20} 76.9 & 85.1 & \cellcolor{gray!20} 87.8  \\
            \cmidrule(r){2-8}
            ECtHR Case Law & \cellcolor{gray!20}72.0 & 75.7  &  \cellcolor{gray!20}\bf 83.3 & 61.9 &  \cellcolor{gray!20}66.3 & 80.1  &  \cellcolor{gray!20}\bf 83.3 \\
            \cmidrule(r){2-8}
            Indian Case Law & \cellcolor{gray!20}65.6 & 70.0 & \cellcolor{gray!20} 65.2 & 56.3 & \cellcolor{gray!20} 58.3 & 73.3 & \cellcolor{gray!20} \bf 76.2 \\
            \cmidrule(r){2-8}
            \textbf{Average} & \cellcolor{gray!20}70.1 & 73.8 & \cellcolor{gray!20} 78.7 & 63.5 & \cellcolor{gray!20} 68.9 & 78.7 & \cellcolor{gray!20} \bf 81.8 \\
            \cmidrule(r){2-8}
            \textbf{Model Rank} & \cellcolor{gray!20}5 & 4 & \cellcolor{gray!20} 2 & 7 & \cellcolor{gray!20} 6 & 2 & \cellcolor{gray!20} 1 \\
    \end{tabular}}
    \caption{Upstream evaluation measured in terms of accuracy (Precision@1) on the Masked Language Modelling (MLM) task across all \textsc{LeXFiles} sub-corpora.}
    \label{tab:mlm_results}
\end{table*}

\section{Experiments}
\label{sec:analysis}

\subsection{Pre-trained Language Models}
\label{sec:models}

We consider 7 large language models to assess their performance with respect to the upstream (MLM), probing, and downstream evaluation:\vspace{2mm}

\noindent\textbf{RoBERTa (Base/Large)} are the original RoBERTa models~\citep{liu-2019-roberta} trained for 64k steps with very large batches on generic corpora; thus do not have any clear legal prior (knowledge).\vspace{2mm}

\noindent\textbf{LegalBERT (Base)} is a legal-oriented BERT model~\citep{devlin-etal-2019-bert} released by \citet{chalkidis-etal-2020-legalbert} trained for 1M steps on legal corpora from EU, UK, CoE, and USA.\vspace{2mm}

\noindent\textbf{CaseLawBERT (Base)} is another legal-oriented BERT released by \citet{zhengguha2021}. CaseLawBERT (which we will refer to as \textit{CL-BERT} henceforth) is trained from scratch for 2M steps on the Harvard Law case
corpus, which comprises 3.4M legal decisions from US federal and state courts.\vspace{2mm}

\noindent\textbf{PoL-BERT (Large)} is a legal-oriented RoBERTa model released by \citet{hendersonkrass2022pileoflaw} trained from scratch for 2M steps on the \textsc{Pile of Law}, a corpus consisting of approx.~256GB of English, mainly US, language legal and administrative text.\vspace{2mm}

\noindent\textbf{LexLM (Base/Large)} are our newly released RoBERTa models. We follow a series of best-practices in language model development:
\begin{enumerate}[label=(\alph*),itemsep=-0.3em]
    \item We warm-start (initialize) our models from the original RoBERTa checkpoints (base or large) of \citet{liu-2019-roberta}.
    \item We train a new tokenizer of 50k BPEs, but we reuse the original embeddings for all lexically overlapping tokens \cite{pfeiffer-etal-2021-unks}.
    \item We continue pre-training our models on the diverse \textsc{LeXFiles} (Section~\ref{sec:corpus}) corpus for additional 1M steps with batches of 512 samples, and a 20/30\% masking rate~\cite{wettig2022should}, for base/large models, respectively. 
    \item We use a sentence sampler with exponential smoothing of the sub-corpora sampling rate following \citet{conneau-etal-2019} since there is a disparate proportion of tokens across sub-corpora (Table~\ref{tab:pile}) and we aim to preserve per-corpus capacity (avoid overfitting).
    \item We consider mixed cased models, similar to all recently developed large PLMs.
\end{enumerate}

\noindent Additional details on LexLM models pre-training can be found in Appendix~\ref{sec:lexlm_details}.

\subsection{Upstream Evaluation}
\label{sec:mlm}

In Table~\ref{tab:mlm_results}, we present the upstream (MLM) performance for all PLMs across the \textsc{LeXFiles} sub-corpora.
The performance is measured in terms of accuracy, i.e. Precision@1 of the masked token to be predicted.
The accuracy is thus averaged over all the masked tokens for each task.
We also provide the average across all tasks, per model.
We observe that results vary across models trained in very different settings (model's capacity, pre-training corpora), while the results also vary across legal sub-corpora.

We want to remind the reader that the upstream evaluation offers a rough idea of a model's capabilities since it relies on random masked sub-words, in which case many of those can be generic and thus highly predictable (e.g. preposition ``of''). This phenomenon further motivates the construction of the \textsc{LegalLAMA} benchmark, in which case only ``legal knowledge sensitive'' words have been masked.

\paragraph{Type of Documents.} In terms of differences across sub-corpora, we observe that the performance on legislation is better compared to case law in 3/4 legal systems, where we have both (EU, UK, US, Canada), with US contractual language being the most predictable for the models which have been trained on it (LexLMs, LegalBERT).

\paragraph{Comparison of PLMs.} Overall, the large LexLM  model outperforms the rest, being 3\% more accurate on average compared to the 2nd best models (base versions of LexLM, and LegalBERT). Such results are expected since LexLMs have been trained in a diverse corpus, similarly to LegalBERT, compared to CL-BERT, and PoL-BERT, which have been trained on US corpora. Over-specialization harms the two US-centric models in a great extend since they are outperformed even from the generic RoBERTa models.

We also observe that LegalBERT outperforms the similarly-sized LexLM in specific sub-corpora (Both EU, UK legislation, ECtHR case law, and US Contracts) that were included in its training. We hypothesize that these results are related to the pre-training data diversity, since LexLMs have been trained in a more diverse corpus including many more documents from different legal systems with a sampling smoothing to preserve capacity per sub-corpus. The larger LexLM model has the capacity to cover all sub-corpora to a greater detail.

\begin{table*}[h!]
    \centering
    \resizebox{\textwidth}{!}{
    \setlength{\tabcolsep}{2.65pt}
    \begin{tabular}{l ccc|c| c| c| c| c| c| c}
 \multicolumn{1}{c}{} & \multicolumn{3}{c}{\textbf{Statistics}} & \multicolumn{7}{c}{\textbf{Models}}\\
 \textbf{Task} & \#T & \#L & \#T/L & \multicolumn{1}{c}{RoBERTa-B} & \multicolumn{1}{c}{RoBERTa-L} & \multicolumn{1}{c}{LegalBERT} & \multicolumn{1}{c}{CL-BERT} & \multicolumn{1}{c}{PoL-BERT} & \multicolumn{1}{c}{LexLM-B} & \multicolumn{1}{c}{LexLM-L}\\
 \cmidrule(lr){2-4}  \cmidrule(lr){5-11}

\textbf{ECHR Articles} & 69 & 13 & 1.0 & \cellcolor{gray!20}39.8 & 41.3 & \cellcolor{gray!20}91.1 & 37.5 & \cellcolor{gray!20}35.2 & 91.4 & \cellcolor{gray!20}\textbf{94.3} \\
\textbf{Contract Sections} & 85 & 20 & 1.3 & \cellcolor{gray!20}23.6 & 44.5 & \cellcolor{gray!20}80.2 & 29.2 & \cellcolor{gray!20}64.8 & \textbf{88.2} & \cellcolor{gray!20}87.3 \\
\textbf{Contract Types} & 150 & 15 & 1.1 & \cellcolor{gray!20}43.4 & 47.8 & \cellcolor{gray!20}82.2 & 54.9 & \cellcolor{gray!20}49.7 & 84.0 & \cellcolor{gray!20}\textbf{86.1} \\
\textbf{Crime Charges (US)} & 118 & 59 & 2.1 & \cellcolor{gray!20}56.3 & 62.4 & \cellcolor{gray!20}51.5 & 62.6 & \cellcolor{gray!20}43.5 & 63.0 & \cellcolor{gray!20}\textbf{68.1} \\
\textbf{Terminology (US)} & 92 & 7 & 2.9 & \cellcolor{gray!20}47.1 & 54.2 & \cellcolor{gray!20}60.5 & 66.7 & \cellcolor{gray!20}44.6 & 66.4 & \cellcolor{gray!20}\textbf{67.5} \\
\textbf{Terminology (EU)} & 164 & 42 & 3.0 & \cellcolor{gray!20}38.0 & 45.3 & \cellcolor{gray!20}63.2 & 38.6 & \cellcolor{gray!20}36.9 & 63.1 & \cellcolor{gray!20}\textbf{70.4} \\ 
\textbf{Terminology (CoE)} & 97 & 250 & 1.2 & \cellcolor{gray!20}45.4 & 53.1 & \cellcolor{gray!20}77.3 & 49.7 & \cellcolor{gray!20}32.8 & 81.3 & \cellcolor{gray!20}\textbf{86.8} \\
\textbf{CC Sections} & 72 & 144 & 2.0 & \cellcolor{gray!20}15.8 & 19.7 & \cellcolor{gray!20}21.9 & 18.4 & \cellcolor{gray!20}19.9 & 50.6 & \cellcolor{gray!20}\textbf{68.8} \\
\cmidrule(lr){2-11}
\multicolumn{1}{c}{} & \multicolumn{3}{r}{\textbf{Average}} & \cellcolor{gray!20}  33.1 & 41.3 & \cellcolor{gray!20}54.8 & 38.0 & \cellcolor{gray!20}36.8 & 70.8 & \cellcolor{gray!20}\textbf{77.4}\\
\cmidrule(lr){2-11}
\multicolumn{1}{c}{} & \multicolumn{3}{r}{\textbf{Model Rank}} & \multicolumn{1}{c|}{7} & \multicolumn{1}{c|}{4} & \multicolumn{1}{c|}{3} & \multicolumn{1}{c|}{5} & \multicolumn{1}{c|}{6} & \multicolumn{1}{c|}{2} & \multicolumn{1}{c}{1} \\

    \end{tabular}
    }
    \caption{The 8 \textsc{LegalLAMA} tasks' statistics regarding the average number of tokens in the input (\#T), the number of labels to predict from (\#L), and the average number of tokens per label (\#T/L) along with the Mean Reciprocal Rank results of the 7 examined PLMs.}
    \label{tab:legal_lama}
    \vspace{-3mm}
\end{table*}

In general, larger models pre-trained on the same corpora (RoBERTas, LexLMs) perform better compared to smaller ones, but in-domain pre-training is a much more important factor for upstream performance, e.g., LegalBERT outperforms RoBERTa-L.

\subsection{Probing Evaluation}
\label{sec:probing}

In Table~\ref{tab:legal_lama}, we present the results across all examined PLMs on \textsc{LegalLAMA}.
We analyze the results from two core perspectives: the prior knowledge and the probing task.

\paragraph{Prior Knowledge.}
The pre-training corpus has a significant impact on the probing performance.
RoBERTa models, having little to no legal prior, were expected to achieve worst performance on all probing tasks.
Surprisingly, CL-BERT and PoL-BERT achieve on-par or sometimes worst performance than RoBERTa (Base \& Large) in most tasks.
Being trained on the ``Harvard Law Case'' corpus (CL-BERT) and the \textsc{Pile of Law} (PoL-BERT), we would have expected better performance than a model without legal prior.
Their pre-training corpora might be lacking diversity, which might cause their poor performance even on Legal-US probing tasks.
LegalBERT (Base), being trained on UK, EU and USA data illustrates important improvement over models without legal prior (RoBERTa) or having only US legal prior (CaseLaw and PoL-BERT).
\textsc{LexLM} models, being trained on the new \textsc{LeXFiles} dataset, show performance improvement over LegalBERT across all tasks, especially on the task of predicting Section Numbers of the Criminal Code of Canada.
Regarding the size of the model, we are able to compare the cased versions of RoBERTa Base/Large and LexLM Base/Large.
As expected, the larger versions offer better performance than the smaller ones on every task.

\definecolor{color_conll}{rgb}{.3,.3,1}
\tikzstyle{comick_style}=[thick, solid, #1, mark size=1.2pt, mark=*, mark options={#1, solid}]
\tikzstyle{mimick_style}=[thick, #1, dashed, mark size=1.2pt, mark=*, mark options={#1, solid}]
\def\height{4.5cm}

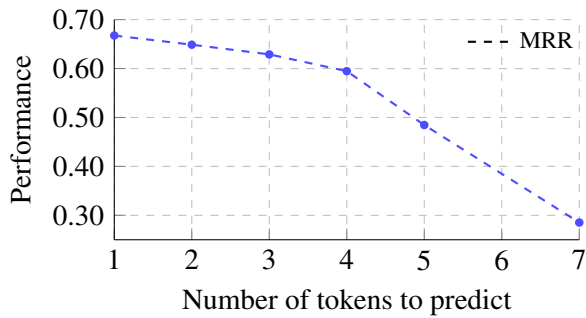
\begin{figure}[h!]
\centering
\begin{tikzpicture}
\def\espacegraphique{45pt}
\def\ymax{12}
\def\xmax{7}
\begin{axis}[
    width=\linewidth, height=\height,
    name=int_cosine,
    grid=major, grid style={dashed,gray!50},
    xlabel=Number of tokens to predict, ylabel=Performance,
    ymin=0.25, ymax=0.7, xmin=1, xmax=\xmax,
    ytick={0.3,0.4,...,0.7},
    xtick={1,2,...,8},
    axis y line*=left, axis x line*=bottom,
    y tick label style={
        /pgf/number format/fixed,
        /pgf/number format/fixed zerofill,
        /pgf/number format/precision=2
    },
    legend pos=north east,
    legend cell align=left,
    legend columns=3,
    transpose legend,
    legend style={draw=none,outer sep=0pt,inner sep=0pt,
    fill=none, font=\footnotesize, yshift=0, xshift=5pt,
        /tikz/column 2/.style={column sep=5pt}
        },
    legend entries={
        MRR,
        }
    ]
    \addlegendimage{no markers, black, dashed, thick}
    \addlegendimage{no markers, black, solid, thick}
    \addplot+[mimick_style=color_conll] table[x=e, y=mrr, col sep=comma]{data/int.csv};

\end{axis}

\end{tikzpicture}
\caption{Models performance on \textsc{LegalLAMA}'s test set with respect to the label complexity. Labels with more than three tokens are much harder to predict. 
}
\label{fig:label_complexity}
\vspace{-3mm}
\end{figure}

\paragraph{Probing Tasks.} We characterize the difficulty of the tasks by their semantic level, the output space (the number of labels to predict from), and the label complexity (how many tokens per label).
We expose the tasks' different characteristics in Table~\ref{tab:legal_lama}.
Given the best-performing model (LexLM-L), we can see that Crime Charges and Legal Terminology (US and EU) are the hardest tasks to solve.
Looking at Table~\ref{tab:legal_lama}, we can see that these three tasks are characterized by a higher label complexity (>2).
We further demonstrate the label complexity impact in Figure~\ref{fig:label_complexity}.
The output space does not seem to have a correlation with the models' performance, since the selected Legal Terminology Topic Clusters (US) has only 7 possible labels, whereas the Criminal Code Section (Canada) has 144 possible labels.
Finally, Crime Charges, being the hardest task to solve, has on average 118 tokens as input and 59 possible labels with moderate complexity, similar to the Terminology tasks (EU and CoE). This suggests that the difficulty of the task is not only driven by the labels' complexity but may rather lie in the lack of contextualization. 
Take for example the following sentence:
\vspace{-1.2mm}
\begin{myquote}{0.1in}
``This case involves perhaps the first prosecution under New York’s new \textbf{[computer crime]} statute, Penal Law article 156, which went into effect on November 1, 1986, just days before the incidents charged herein.''
\end{myquote}
\vspace{-1.2mm}
The only contextual hint the PLMs have to predict the correct tokens (\textbf{[computer crime]}) is the utterance ``Penal Law article 156, which went into effect on November 1, 1986''. This is the opposite task of predicting article numbers given a context, which is much more difficult than predicting the actual context because the output space is larger.\footnote{The actual tokens predicted by the best-performing examined PLM were ``sexual'' and ``abuse''.}

 \begin{table*}[h!]
    \centering
    \resizebox{\textwidth}{!}{
    \begin{tabular}{l cc| cc| cc| cc| cc| cc| cc}
 & \multicolumn{2}{c}{RoBERTa-B} & \multicolumn{2}{c}{RoBERTa-L} & \multicolumn{2}{c}{LegalBERT} & \multicolumn{2}{c}{CL-BERT} & \multicolumn{2}{c}{PoL-BERT} & \multicolumn{2}{c}{LexLM-B} & \multicolumn{2}{c}{LexLM-L}\\
 \cmidrule(lr){2-15}
 \bf Task & \microf & \macrof & \microf & \macrof & \microf & \macrof & \microf & \macrof & \microf & \macrof & \microf & \macrof & \microf & \macrof \\
\cmidrule(lr){2-3} \cmidrule(lr){4-5} \cmidrule(lr){6-7} \cmidrule(lr){8-9} \cmidrule(lr){10-11} \cmidrule(lr){12-13} \cmidrule(lr){14-15} 
\bf ECtHR & \cellcolor{gray!20} 61.2 & 40.5 & \cellcolor{gray!20} 74.2 & 51.5 & \cellcolor{gray!20} 59.1 & 37.2 & \cellcolor{gray!20} 53.6 & 29.1 & \cellcolor{gray!20} 69.1 & 46.9 & \cellcolor{gray!20} 63.2 & 41.8 & \cellcolor{gray!20} \bf 76.7 & \bf 57.9 \\
\bf LEDGAR & \cellcolor{gray!20} 80.5 & 62.6 & \cellcolor{gray!20} 83.6 & 71.5 & \cellcolor{gray!20} 81.2 & 64.7 & \cellcolor{gray!20} 80.9 & 64.0 & \cellcolor{gray!20} 83.3 & 71.4 & \cellcolor{gray!20} 82.5 & 66.8 & \cellcolor{gray!20} \bf 84.7 & \bf 72.8 \\
\bf CNLI & \cellcolor{gray!20} 66.8 & 48.6 & \cellcolor{gray!20} 68.0 & 63.5 & \cellcolor{gray!20} \bf 70.2 & \bf 65.6 & \cellcolor{gray!20} 69.0 & 64.6 & \cellcolor{gray!20} 68.3 & 64.1 & \cellcolor{gray!20} 61.6 & 42.9 & \cellcolor{gray!20} 69.7 & 64.5 \\
\bf SCOTUS & \cellcolor{gray!20} 65.0 & 36.0 & \cellcolor{gray!20} 68.9 & 41.4 & \cellcolor{gray!20} 60.9 & 31.2 & \cellcolor{gray!20} 62.9 & 33.8 & \cellcolor{gray!20} 66.3 & 39.5 & \cellcolor{gray!20} 66.9 & 37.7 & \cellcolor{gray!20} \bf 71.1 & \bf 43.9\\
\bf CaseHOLD & \cellcolor{gray!20} 72.7 & 72.7 & \cellcolor{gray!20} 75.6 & 75.6 & \cellcolor{gray!20} 76.1 & 76.1 & \cellcolor{gray!20} 77.6 & 77.6 & \cellcolor{gray!20} 73.7 & 73.7 & \cellcolor{gray!20} 74.8 & 74.8 & \cellcolor{gray!20} \bf 78.5 & \bf 78.5  \\
\bf EURLEX & \cellcolor{gray!20} 33.4 & 06.1 & \cellcolor{gray!20} 62.7 & 27.1 & \cellcolor{gray!20} 27.7 & 04.0 & \cellcolor{gray!20} 27.0 & 04.7  & \cellcolor{gray!20} 60.5 & 25.4 & \cellcolor{gray!20} 34.2 & 06.9 & \cellcolor{gray!20} \bf 63.1 & \bf 28.0 \\
\cmidrule(lr){1-15}
\textbf{Average} & \cellcolor{gray!20} 58.4 & 22.5 & \cellcolor{gray!20} 71.5 & 48.6 & \cellcolor{gray!20} 55.0 & 17.1 & \cellcolor{gray!20} 53.9 & 18.7 & \cellcolor{gray!20} 69.5 & 46.4 & \cellcolor{gray!20} 59.0 & 24.3  & \cellcolor{gray!20} \bf 73.3 & \bf 51.0  \\
\cmidrule(lr){1-15}
\textbf{Upstream} & \multicolumn{2}{c|}{5} & \multicolumn{2}{c|}{4} & \multicolumn{2}{c|}{2} & \multicolumn{2}{c|}{7} & \multicolumn{2}{c|}{6} & \multicolumn{2}{c|}{2} & \multicolumn{2}{c}{1} \\
\textbf{Probing} & \multicolumn{2}{c|}{7} & \multicolumn{2}{c|}{4} & \multicolumn{2}{c|}{3} & \multicolumn{2}{c|}{5} & \multicolumn{2}{c|}{6} & \multicolumn{2}{c|}{2} & \multicolumn{2}{c}{1} \\
\textbf{Downstream} & \multicolumn{2}{c|}{5} & \multicolumn{2}{c|}{2} & \multicolumn{2}{c|}{6} & \multicolumn{2}{c|}{7} & \multicolumn{2}{c|}{3} & \multicolumn{2}{c|}{4} & \multicolumn{2}{c}{1} \\
\end{tabular}
}
 \caption{Test Results for all models across all downstream tasks after fine-tuning for a single epoch.}
    \label{tab:donwstream}
\end{table*}

\subsection{Downstream Evaluation}
\label{sec:downstream}

For downstream evaluation, we conduct experiments for 6 legal classification tasks, 5 part of \textsc{LexGLUE}~\cite{chalkidis-etal-2022-lexglue}, covering US contracts, US, EU, and ECHR law.\vspace{2mm}

\noindent\textbf{ECtHR (Task B)}~\cite{chalkidis-etal-2021-paragraph} is a multi-label topic classification task, where given the facts of an ECtHR case, the model has to predict the alleged violated ECHR article among 10 such articles (e.g., ``Art 3. - Prohibition of Torture'', ``Art. 6 - Right to Fair Trial'').\vspace{2mm} 

\noindent\textbf{LEDGAR}~\cite{tuggener-etal-2020-ledgar} is a single-label multi-class topic classification task, where given a contractual paragraph, the model has to predict one of the correct topic among 100 topics (e.g., ``Limitation of Liability'', ``Arbitration'').\vspace{2mm}

\noindent\textbf{ContractNLI}~\cite{koreeda-manning-2021-contractnli-dataset} is a contract-based Natural Language Inference (NLI) task, where given an Non-Disclosure Agreement (NDA) and one out 17 templated \emph{hypotheses} (e.g., ``The Party may share some Confidential Information with some third-parties.''), the model has to predict if the hypothesis is (\emph{entailed}, \emph{contradicted}, or is \emph{neutral}) to the terms of the NDA.\vspace{2mm} 

\noindent\textbf{SCOTUS}~\cite{chalkidis-etal-2022-lexglue} is a single-label multi-class topic classification task, where given a Supreme Court of US (SCOTUS) opinion, the model has to predict the relevant area among 14 issue areas (e.g., ``Civil Rights'', ``Judicial Power'').\vspace{2mm}

\noindent\textbf{CaseHOLD}~\cite{zhengguha2021} is a multiple choice QA classification task, where given a paragraph from a US legal opinion where a legal rule (holding) is masked, the model has to predict the applicable rule among 5 alternatives (the correct one and 2 irrelevant presented in other cases).\vspace{2mm}

\noindent\textbf{EURLEX}~\cite{chalkidis-etal-2021-multieurlex} is a multi-label topic classification task, where given an EU law, the model has to predict the correct EUROVOC concept among hundred concepts (e.g., ``Environmental Policy'', ``International Trade'').\vspace{2mm}\vspace{2mm}

\begin{figure}[t!]
\centering
\resizebox{0.95\columnwidth}{!}{
\begin{tikzpicture}
\def\espacegraphique{45pt}
\def\ymax{12}
\def\xmax{6}
\begin{axis}[
    width=\columnwidth, height=\height,
    name=int_micro,
    grid=major, grid style={dashed,gray!50},
    xlabel=Training Epochs, ylabel=Performance (\microf),
    ymin=73, ymax=82, xmin=1, xmax=5,
    ytick={70,75,...,85},
    xtick={1,2,...,5},
    axis y line*=left, axis x line*=bottom,
    legend pos=south east,
    legend cell align=left,
    legend columns=3,
    transpose legend,
    legend style={draw=none,outer sep=0pt,inner sep=0pt,
    fill=none, font=\footnotesize, yshift=0, xshift=5pt,
        /tikz/column 2/.style={column sep=5pt}
        },
    legend entries={
        LexLM-L,
        RoBERTa-L
        }
    ]
    \addlegendimage{no markers, black, dashed, thick}
    \addlegendimage{no markers, black, solid, thick}

    \addplot+[comick_style=color_conll] table[x=e, y=roberta, col sep=comma]{data/micro.csv};

    \addplot+[mimick_style=color_conll] table[x=e, y=lexlm, col sep=comma]{data/micro.csv};

\end{axis}
\end{tikzpicture}
}
\caption{Development Results of RoBERTa and LexLM large on ECtHR across 5 training epochs.}
\label{fig:ecthr_5_epochs}
\vspace{-3mm}
\end{figure}
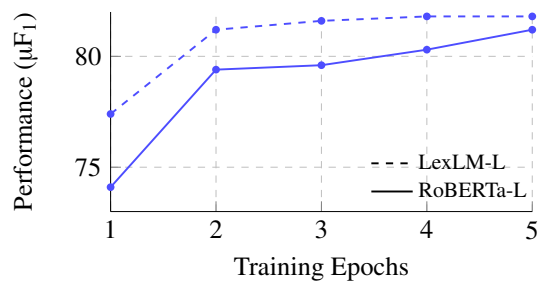

\noindent  We fine-tune all examined PLMs (Section~\ref{sec:models}) for a single epoch with a learning rate of $1e\!-\!5$ leading to a small number of updates. We are interested to examine how fast each model convergence based on its prior knowledge; in other words, what can a model learn in a single pass over training data? Finetuning models for many epochs over large datasets will eventually lead to a full re-parameterization of the models, in which case the importance of prior knowledge will diminish compromise the goal of our study (Figure~\ref{fig:ecthr_5_epochs}).\footnote{In most tasks, models fully converge after approx.~5 epochs with improved performance, and the relative differences between generic and legal-oriented models are diminished~\cite{chalkidis-etal-2022-lexglue}.}

For all tasks, we use standard N-way classifiers with a classification head~\cite{devlin-etal-2019-bert}. For ECtHR, and SCOTUS, involving long documents, we warm-start Longformer~\cite{Longformer} models from each PLM's parameters to encode up to 2048 tokens. We evaluate classification performance with micro-F1 (\microf) and macro-F1 (\macrof) across tasks following \citet{chalkidis-etal-2022-lexglue}.

\paragraph{Results}
In Table~\ref{tab:donwstream}, we present the test results across all tasks/datasets. We analyze the results from two perspectives: model's capacity (size), and prior legal knowledge abducted via pre-training.\vspace{2mm}

\noindent\textbf{Model's capacity (size)} strongly correlates with the overall downstream performance. Across all tasks, there are 2/6 exceptions (CNLI and CaseHOLD) where LegalBERT outperforms larger PLMs. Both tasks are using sentence pairs, a setup used in BERT's pre-training, but not in RoBERTa, which may bring LegalBERT, a BERT-based model, in a better initial condition co-considering the minimal updates steps, compared to all large models following the RoBERTa pre-training setup, which do no use pairs of sentences or optimized based on a sentence-level objective (NSP).\vspace{2mm}

\noindent\textbf{Legal Knowledge} also plays an important role following the model's capacity (size). We observe that LexLM-B trained in the diverse \textsc{LeXFiles} corpus outperforms the equally-sized RoBERTa-B model in 5/6 tasks, while LegalBERT and CL-BERT outperform it only in 3 out of 6 tasks. In this case, the results are mixed, i.e., acquaintance of legal knowledge as expressed by upstream (Section~\ref{sec:mlm}) and probing (Section~\ref{sec:probing}) performance does not correlate with downstream performance.

In the case of large-sized models, LexLM-L outperform RoBERTa-L across all tasks, while PoL-BERT trained on the US-biased \textsc{Pile of Law} corpus is outperformed by RoBERTa-L in 5 out of 6 tasks. Given the results with respect to upstream and probing performance, RoBERTa-L has a better legal prior; so in these regards, acquaintance of legal knowledge fully correlates with downstream performance in the large models' regime.

\section{Release of Resources}
We release our code base to assure reproducibility and let others extend our study by experimenting with other PLMs, or develop new ones.\footnote{\url{https://github.com/coastalcph/lexlms}} The new LexLM models (Section~\ref{sec:models}), the \textsc{LeXFiles} corpus~\footnote{\url{https://huggingface.co/datasets/lexlms/lex_files}} (Section~\ref{sec:corpus}), and the \textsc{LegalLAMA} benchmark~\footnote{\url{https://huggingface.co/datasets/lexlms/legal_lama}} (Section~\ref{sec:probing}) are available on Hugging Face Hub~\cite{lhoest2021datasets}.\footnote{\url{https://huggingface.co/lexlms}}

\section{Conclusions and Future Work}
In this work, we introduced a multinational English legal corpus (\textsc{LeXFiles}) and a legal knowledge probing benchmark (\textsc{LegalLAMA}) to facilitate training and detailed analysis of legal-oriented PLMs. We also released two new legal PLMs and evaluate them alongside others on \textsc{LegalLAMA} and \textsc{LexGLUE}. Based on our analysis (Section~\ref{sec:analysis}), we make the following general observations:
\begin{enumerate}[label=(\alph*),itemsep=-0.3em]
    \item The use of diverse legal corpora leads to better overall upstream performance (Section~\ref{sec:mlm}).
    \item We find that probing performance strongly correlates with upstream performance in related legal topics (Section~\ref{sec:probing}).
    \item For both upstream, and probing performance, the selection of pre-training corpora has a much larger effect compared to model's capacity (Sections~\ref{sec:mlm}-\ref{sec:probing}). Nonetheless, larger models pre-trained on similar corpora have better overall performance.
    \item Downstream performance is mainly driven by the model's capacity and prior legal knowledge which can be estimated by upstream and probing performance (Section~\ref{sec:downstream}).
\end{enumerate}

In future work, we plan to further analyze the learning dynamics of legal language models by comparing their representations with representations derived from legal knowledge bases. Given the availability of the new resources, the development of instruction-following~\cite{wei-etal-2022} fine-tuned legal-oriented GPT-like~\cite{instructgpt} models is also an anticipated direction.

\section*{Limitations}
\paragraph{Diversity of Corpora} While the newly introduced \textsc{LeXFiles} corpus is significantly more diverse compared to the \textsc{Pile of Law} corpus of \citet{hendersonkrass2022pileoflaw}, it is still an English-only corpus covering only 6 legal systems (EU, UK, CoE, US, India, Canada). Despite, the fact that we can train better models (LexLMs) and evaluate these models across these corpora, in future work, we should extend our analysis to cover even more languages and legal systems, and a higher granularity in the labeling of legal fields within these systems. Not only will this help support the inclusion of other legal traditions but also adding more linguistic and cultural diversity will help us better understand the robustness of existing methods.

Similarly, the newly introduced \textsc{LegalLAMA} benchmark consists of 8 sub-tasks targeting EU, ECHR, US, and Canadian jurisdictions in a very controlled setting; where examples were automatically extracted. While on this benchmark, legal-oriented PLMs has demonstrated a significant degree of ``understanding" of legal language and legal topics, this benchmark should be further expanded with more sub-tasks to evaluate the acquaintance of legal knowledge across more legal systems and topics, and possibly cleansed from both very easy and unsolvable examples.

\paragraph{Model Considerations} In this work, we consider encoder-only (BERT-like) models up to approx. 350M parameters, while recent work on the development of Large Language Models (LLMs) \cite{kaplan2020, brown-etal-gpt3, hoffman-etal-chinchilla,Chowdhery2022} is mainly targeting billion-parameter-sized models (10-100Bs of parameters) that usually follow a decoder-only, e.g., GPT~\cite{Radford2018ImprovingLU}, or encoder-decoder, e.g., T5~\cite{raffel2020}, architecture.  Moreover, new paradigms of training PLMs have been introduced, such as \emph{instruction-based finetuning}~\cite{wei-etal-2022}, and \emph{alignment} via Reinforcement Learning from Human Feedback  (RLHF)~\cite{Stiennon2020, instructgpt}. Latest GPT models~\cite{instructgpt} have recently shown significant zero-shot progress on law-related tasks such as bar examination question answering \cite{katz2023gpt}. Thus, future work should follow the most recent advances by pre-training much larger auto-regressive GTP-like models that seem to lead to emergent zero-shot and few-shot capabilities. 

\paragraph{Evaluation Considerations}
In Section~\ref{sec:legallama}, we present how we account for and evaluate multi-token expressions (terms) on the \textsc{LegalLAMA} benchmark; we are open to ideas on how we should possibly improve the current approach to provide a fairer and more robust evaluation framework across all models. Similarly, in Section~\ref{sec:downstream}, we fine-tune all examined PLMs for a single epoch to avoid extreme over-reparameterization and better estimate how model's knowledge affects convergence and performance. Nonetheless, there are possibly better approaches to control for these aspects, e.g., Adapter-based~\cite{ruckle-etal-2021-adapterdrop} finetuning, or other approaches, such as LoRA~\cite{hu2022lora}.

\paragraph{Beyond Performance} While we consider a multi-facet analysis, we do not cover other interesting dimensions that should also be explored, especially since law is a very sensitive application domain; for instance trustworthiness-related topics, such as model interpretability~\cite{chalkidis-etal-2021-paragraph, malik-etal-2021-ildc}, and fairness~\cite{chalkidis-etal-2022-fairlex}.  Future work can build from the results reported herein to explore these important topics. 

\section*{Ethics Statement}
The scope of this work is to examine the performance of legal-oriented PLMs from a multi-facet perspective and broaden the discussion to help practitioners build assisting technology for legal professionals and laypersons. We believe that this is an important application field, where research should be conducted \cite{tsarapatsanis-aletras-2021-ethical} to improve legal services and democratize law, while also highlighting (informing the audience on) the various multi-aspect shortcomings seeking a responsible and ethical (fair) deployment of legal-oriented technologies. 

In this direction, we introduce new resources covering various legal systems to build new models that better represent law and better assess their capabilities. All newly developed and published resources are based on publicly available data, most of them scattered on several web portals.

\section*{Acknowledgments}
This work was partly funded by the Innovation Fund Denmark (IFD, \url{https://innovationsfonden.dk/en}) and the Fonds de recherche du Qu\'ebec -- Nature et technologies (FRQNT, \url{https://frq.gouv.qc.ca/nature-et-technologies/}).

\bibliography{anthology,custom}

\begin{thebibliography}{38}
\expandafter\ifx\csname natexlab\endcsname\relax\def\natexlab#1{#1}\fi

\bibitem[{Beltagy et~al.(2020)Beltagy, Peters, and Cohan}]{Longformer}
Iz~Beltagy, Matthew~E. Peters, and Arman Cohan. 2020.
\newblock \href {https://arxiv.org/abs/2004.05150} {Longformer: The
  long-document transformer}.
\newblock \emph{CoRR}, abs/2004.05150.

\bibitem[{Brown et~al.(2020)Brown, Mann, Ryder, Subbiah, Kaplan, Dhariwal,
  Neelakantan, Shyam, Sastry, Askell, Agarwal, Herbert-Voss, Krueger, Henighan,
  Child, Ramesh, Ziegler, Wu, Winter, Hesse, Chen, Sigler, Litwin, Gray, Chess,
  Clark, Berner, McCandlish, Radford, Sutskever, and Amodei}]{brown-etal-gpt3}
Tom Brown, Benjamin Mann, Nick Ryder, Melanie Subbiah, Jared~D Kaplan, Prafulla
  Dhariwal, Arvind Neelakantan, Pranav Shyam, Girish Sastry, Amanda Askell,
  Sandhini Agarwal, Ariel Herbert-Voss, Gretchen Krueger, Tom Henighan, Rewon
  Child, Aditya Ramesh, Daniel Ziegler, Jeffrey Wu, Clemens Winter, Chris
  Hesse, Mark Chen, Eric Sigler, Mateusz Litwin, Scott Gray, Benjamin Chess,
  Jack Clark, Christopher Berner, Sam McCandlish, Alec Radford, Ilya Sutskever,
  and Dario Amodei. 2020.
\newblock \href
  {https://proceedings.neurips.cc/paper/2020/file/1457c0d6bfcb4967418bfb8ac142f64a-Paper.pdf}
  {Language models are few-shot learners}.
\newblock In \emph{Advances in Neural Information Processing Systems},
  volume~33, pages 1877--1901. Curran Associates, Inc.

\bibitem[{Chalkidis et~al.(2021{\natexlab{a}})Chalkidis, Fergadiotis, and
  Androutsopoulos}]{chalkidis-etal-2021-multieurlex}
Ilias Chalkidis, Manos Fergadiotis, and Ion Androutsopoulos.
  2021{\natexlab{a}}.
\newblock \href {https://doi.org/10.18653/v1/2021.emnlp-main.559}
  {{M}ulti{EURLEX} - a multi-lingual and multi-label legal document
  classification dataset for zero-shot cross-lingual transfer}.
\newblock In \emph{Proceedings of the 2021 Conference on Empirical Methods in
  Natural Language Processing}, pages 6974--6996, Online and Punta Cana,
  Dominican Republic. Association for Computational Linguistics.

\bibitem[{Chalkidis et~al.(2020)Chalkidis, Fergadiotis, Malakasiotis, Aletras,
  and Androutsopoulos}]{chalkidis-etal-2020-legalbert}
Ilias Chalkidis, Manos Fergadiotis, Prodromos Malakasiotis, Nikolaos Aletras,
  and Ion Androutsopoulos. 2020.
\newblock \href {https://doi.org/10.18653/v1/2020.findings-emnlp.261}
  {{LEGAL}-{BERT}: The muppets straight out of law school}.
\newblock In \emph{Findings of the Association for Computational Linguistics:
  EMNLP 2020}, pages 2898--2904, Online.

\bibitem[{Chalkidis et~al.(2021{\natexlab{b}})Chalkidis, Fergadiotis,
  Tsarapatsanis, Aletras, Androutsopoulos, and
  Malakasiotis}]{chalkidis-etal-2021-paragraph}
Ilias Chalkidis, Manos Fergadiotis, Dimitrios Tsarapatsanis, Nikolaos Aletras,
  Ion Androutsopoulos, and Prodromos Malakasiotis. 2021{\natexlab{b}}.
\newblock \href {https://doi.org/10.18653/v1/2021.naacl-main.22}
  {Paragraph-level rationale extraction through regularization: A case study on
  {E}uropean court of human rights cases}.
\newblock In \emph{Proceedings of the 2021 Conference of the North American
  Chapter of the Association for Computational Linguistics: Human Language
  Technologies}, pages 226--241, Online. Association for Computational
  Linguistics.

\bibitem[{Chalkidis et~al.(2022{\natexlab{a}})Chalkidis, Jana, Hartung,
  Bommarito, Androutsopoulos, Katz, and Aletras}]{chalkidis-etal-2022-lexglue}
Ilias Chalkidis, Abhik Jana, Dirk Hartung, Michael Bommarito, Ion
  Androutsopoulos, Daniel Katz, and Nikolaos Aletras. 2022{\natexlab{a}}.
\newblock \href {https://doi.org/10.18653/v1/2022.acl-long.297} {{L}ex{GLUE}: A
  benchmark dataset for legal language understanding in {E}nglish}.
\newblock In \emph{Proceedings of the 60th Annual Meeting of the Association
  for Computational Linguistics (Volume 1: Long Papers)}, pages 4310--4330,
  Dublin, Ireland. Association for Computational Linguistics.

\bibitem[{Chalkidis et~al.(2022{\natexlab{b}})Chalkidis, Pasini, Zhang, Tomada,
  Schwemer, and S{\o}gaard}]{chalkidis-etal-2022-fairlex}
Ilias Chalkidis, Tommaso Pasini, Sheng Zhang, Letizia Tomada, Sebastian
  Schwemer, and Anders S{\o}gaard. 2022{\natexlab{b}}.
\newblock \href {https://doi.org/10.18653/v1/2022.acl-long.301} {{F}air{L}ex: A
  multilingual benchmark for evaluating fairness in legal text processing}.
\newblock In \emph{Proceedings of the 60th Annual Meeting of the Association
  for Computational Linguistics (Volume 1: Long Papers)}, pages 4389--4406,
  Dublin, Ireland. Association for Computational Linguistics.

\bibitem[{Chowdhery et~al.(2022)Chowdhery, Narang, Devlin, Bosma, Mishra,
  Roberts, Barham, Chung, Sutton, Gehrmann, Schuh, Shi, Tsvyashchenko, Maynez,
  Rao, Barnes, Tay, Shazeer, Prabhakaran, Reif, Du, Hutchinson, Pope, Bradbury,
  Austin, Isard, Gur-Ari, Yin, Duke, Levskaya, Ghemawat, Dev, Michalewski,
  Garcia, Misra, Robinson, Fedus, Zhou, Ippolito, Luan, Lim, Zoph, Spiridonov,
  Sepassi, Dohan, Agrawal, Omernick, Dai, Pillai, Pellat, Lewkowycz, Moreira,
  Child, Polozov, Lee, Zhou, Wang, Saeta, Diaz, Firat, Catasta, Wei,
  Meier-Hellstern, Eck, Dean, Petrov, and Fiedel}]{Chowdhery2022}
Aakanksha Chowdhery, Sharan Narang, Jacob Devlin, Maarten Bosma, Gaurav Mishra,
  Adam Roberts, Paul Barham, Hyung~Won Chung, Charles Sutton, Sebastian
  Gehrmann, Parker Schuh, Kensen Shi, Sasha Tsvyashchenko, Joshua Maynez,
  Abhishek Rao, Parker Barnes, Yi~Tay, Noam Shazeer, Vinodkumar Prabhakaran,
  Emily Reif, Nan Du, Ben Hutchinson, Reiner Pope, James Bradbury, Jacob
  Austin, Michael Isard, Guy Gur-Ari, Pengcheng Yin, Toju Duke, Anselm
  Levskaya, Sanjay Ghemawat, Sunipa Dev, Henryk Michalewski, Xavier Garcia,
  Vedant Misra, Kevin Robinson, Liam Fedus, Denny Zhou, Daphne Ippolito, David
  Luan, Hyeontaek Lim, Barret Zoph, Alexander Spiridonov, Ryan Sepassi, David
  Dohan, Shivani Agrawal, Mark Omernick, Andrew~M. Dai,
  Thanumalayan~Sankaranarayana Pillai, Marie Pellat, Aitor Lewkowycz, Erica
  Moreira, Rewon Child, Oleksandr Polozov, Katherine Lee, Zongwei Zhou, Xuezhi
  Wang, Brennan Saeta, Mark Diaz, Orhan Firat, Michele Catasta, Jason Wei,
  Kathy Meier-Hellstern, Douglas Eck, Jeff Dean, Slav Petrov, and Noah Fiedel.
  2022.
\newblock \href {https://doi.org/10.48550/ARXIV.2204.02311} {Palm: Scaling
  language modeling with pathways}.

\bibitem[{Conneau et~al.(2019)Conneau, Khandelwal, Goyal, Chaudhary, Wenzek,
  Guzm{\'{a}}n, Grave, Ott, Zettlemoyer, and Stoyanov}]{conneau-etal-2019}
Alexis Conneau, Kartikay Khandelwal, Naman Goyal, Vishrav Chaudhary, Guillaume
  Wenzek, Francisco Guzm{\'{a}}n, Edouard Grave, Myle Ott, Luke Zettlemoyer,
  and Veselin Stoyanov. 2019.
\newblock \href {http://arxiv.org/abs/1911.02116} {Unsupervised cross-lingual
  representation learning at scale}.
\newblock \emph{CoRR}, abs/1911.02116.

\bibitem[{Devlin et~al.(2019)Devlin, Chang, Lee, and
  Toutanova}]{devlin-etal-2019-bert}
Jacob Devlin, Ming-Wei Chang, Kenton Lee, and Kristina Toutanova. 2019.
\newblock \href {https://doi.org/10.18653/v1/N19-1423} {{BERT}: Pre-training of
  deep bidirectional transformers for language understanding}.
\newblock In \emph{Proceedings of the 2019 Conference of the North {A}merican
  Chapter of the Association for Computational Linguistics: Human Language
  Technologies, Volume 1 (Long and Short Papers)}, pages 4171--4186,
  Minneapolis, Minnesota. Association for Computational Linguistics.

\bibitem[{Friedman and Hayden(2017)}]{Friedman2017}
Lawrence~M. Friedman and Grant~M. Hayden. 2017.
\newblock \href {https://doi.org/10.1093/acprof:oso/9780190460587.003.0001}
  {{1What Is a Legal System?}}
\newblock In \emph{{American Law: An Introduction}}. Oxford University Press.

\bibitem[{Habernal et~al.(2022)Habernal, Faber, Recchia, Bretthauer, Gurevych,
  Spiecker~genannt Döhmann, and Burchard}]{habernal-etal-2022-argument}
Ivan Habernal, Daniel Faber, Nicola Recchia, Sebastian Bretthauer, Iryna
  Gurevych, Indra Spiecker~genannt Döhmann, and Christoph Burchard. 2022.
\newblock \href {https://doi.org/10.48550/arXiv.2208.06178} {{Mining Legal
  Arguments in Court Decisions}}.
\newblock \emph{arXiv preprint}.

\bibitem[{Henderson* et~al.(2022)Henderson*, Krass*, Zheng, Guha, Manning,
  Jurafsky, and Ho}]{hendersonkrass2022pileoflaw}
Peter Henderson*, Mark~S. Krass*, Lucia Zheng, Neel Guha, Christopher~D.
  Manning, Dan Jurafsky, and Daniel~E. Ho. 2022.
\newblock \href {https://arxiv.org/abs/2207.00220} {Pile of law: Learning
  responsible data filtering from the law and a 256gb open-source legal
  dataset}.

\bibitem[{Hoffmann et~al.(2022)Hoffmann, Borgeaud, Mensch, Buchatskaya, Cai,
  Rutherford, Casas, Hendricks, Welbl, Clark, Hennigan, Noland, Millican,
  Driessche, Damoc, Guy, Osindero, Simonyan, Elsen, Rae, Vinyals, and
  Sifre}]{hoffman-etal-chinchilla}
Jordan Hoffmann, Sebastian Borgeaud, Arthur Mensch, Elena Buchatskaya, Trevor
  Cai, Eliza Rutherford, Diego de~Las Casas, Lisa~Anne Hendricks, Johannes
  Welbl, Aidan Clark, Tom Hennigan, Eric Noland, Katie Millican, George van~den
  Driessche, Bogdan Damoc, Aurelia Guy, Simon Osindero, Karen Simonyan, Erich
  Elsen, Jack~W. Rae, Oriol Vinyals, and Laurent Sifre. 2022.
\newblock \href {https://doi.org/10.48550/ARXIV.2203.15556} {Training
  compute-optimal large language models}.

\bibitem[{Hu et~al.(2022)Hu, yelong shen, Wallis, Allen-Zhu, Li, Wang, Wang,
  and Chen}]{hu2022lora}
Edward~J Hu, yelong shen, Phillip Wallis, Zeyuan Allen-Zhu, Yuanzhi Li, Shean
  Wang, Lu~Wang, and Weizhu Chen. 2022.
\newblock \href {https://openreview.net/forum?id=nZeVKeeFYf9} {Lo{RA}: Low-rank
  adaptation of large language models}.
\newblock In \emph{International Conference on Learning Representations}.

\bibitem[{Hwang et~al.(2022)Hwang, Lee, Cho, Lee, and Seo}]{hwang2022a}
Wonseok Hwang, Dongjun Lee, Kyoungyeon Cho, Hanuhl Lee, and Minjoon Seo. 2022.
\newblock \href {https://openreview.net/forum?id=TaARsI_Iio} {A multi-task
  benchmark for korean legal language understanding and judgement prediction}.
\newblock In \emph{Thirty-sixth Conference on Neural Information Processing
  Systems Datasets and Benchmarks Track}.

\bibitem[{Kaplan et~al.(2020)Kaplan, McCandlish, Henighan, Brown, Chess, Child,
  Gray, Radford, Wu, and Amodei}]{kaplan2020}
Jared Kaplan, Sam McCandlish, Tom Henighan, Tom~B. Brown, Benjamin Chess, Rewon
  Child, Scott Gray, Alec Radford, Jeffrey Wu, and Dario Amodei. 2020.
\newblock \href {http://arxiv.org/abs/2001.08361} {Scaling laws for neural
  language models}.
\newblock \emph{CoRR}, abs/2001.08361.

\bibitem[{Katz et~al.(2023)Katz, Bommarito, Gao, and Arredondo}]{katz2023gpt}
Daniel~Martin Katz, Michael~James Bommarito, Shang Gao, and Pablo Arredondo.
  2023.
\newblock \href {https://papers.ssrn.com/sol3/papers.cfm?abstract_id=4389233}
  {Gpt-4 passes the bar exam}.

\bibitem[{Koreeda and Manning(2021)}]{koreeda-manning-2021-contractnli-dataset}
Yuta Koreeda and Christopher Manning. 2021.
\newblock \href {https://doi.org/10.18653/v1/2021.findings-emnlp.164}
  {{C}ontract{NLI}: A dataset for document-level natural language inference for
  contracts}.
\newblock In \emph{Findings of the Association for Computational Linguistics:
  EMNLP 2021}, pages 1907--1919, Punta Cana, Dominican Republic. Association
  for Computational Linguistics.

\bibitem[{Lhoest et~al.(2021)Lhoest, del Moral, Jernite, Thakur, von Platen,
  Patil, Chaumond, Drame, Plu, Tunstall, Davison, Šaško, Chhablani, Malik,
  Brandeis, Scao, Sanh, Xu, Patry, McMillan-Major, Schmid, Gugger, Delangue,
  Matussière, Debut, Bekman, Cistac, Goehringer, Mustar, Lagunas, Rush, and
  Wolf}]{lhoest2021datasets}
Quentin Lhoest, Albert~Villanova del Moral, Yacine Jernite, Abhishek Thakur,
  Patrick von Platen, Suraj Patil, Julien Chaumond, Mariama Drame, Julien Plu,
  Lewis Tunstall, Joe Davison, Mario Šaško, Gunjan Chhablani, Bhavitvya
  Malik, Simon Brandeis, Teven~Le Scao, Victor Sanh, Canwen Xu, Nicolas Patry,
  Angelina McMillan-Major, Philipp Schmid, Sylvain Gugger, Clément Delangue,
  Théo Matussière, Lysandre Debut, Stas Bekman, Pierric Cistac, Thibault
  Goehringer, Victor Mustar, François Lagunas, Alexander~M. Rush, and Thomas
  Wolf. 2021.
\newblock \href {http://arxiv.org/abs/2109.02846} {Datasets: A community
  library for natural language processing}.

\bibitem[{Liu et~al.(2019)Liu, Ott, Goyal, Du, Joshi, Chen, Levy, Lewis,
  Zettlemoyer, and Stoyanov}]{liu-2019-roberta}
Yinhan Liu, Myle Ott, Naman Goyal, Jingfei Du, Mandar Joshi, Danqi Chen, Omer
  Levy, Mike Lewis, Luke Zettlemoyer, and Veselin Stoyanov. 2019.
\newblock \href {https://arxiv.org/abs/1907.11692} {Roberta: A robustly
  optimized bert pretraining approach}.
\newblock \emph{arXiv preprint arXiv:1907.11692}.

\bibitem[{Malik et~al.(2021)Malik, Sanjay, Nigam, Ghosh, Guha, Bhattacharya,
  and Modi}]{malik-etal-2021-ildc}
Vijit Malik, Rishabh Sanjay, Shubham~Kumar Nigam, Kripabandhu Ghosh,
  Shouvik~Kumar Guha, Arnab Bhattacharya, and Ashutosh Modi. 2021.
\newblock \href {https://doi.org/10.18653/v1/2021.acl-long.313} {{ILDC} for
  {CJPE}: {I}ndian legal documents corpus for court judgment prediction and
  explanation}.
\newblock In \emph{Proceedings of the 59th Annual Meeting of the Association
  for Computational Linguistics and the 11th International Joint Conference on
  Natural Language Processing (Volume 1: Long Papers)}, pages 4046--4062,
  Online. Association for Computational Linguistics.

\bibitem[{Niklaus et~al.(2023)Niklaus, Matoshi, Rani, Galassi, Stürmer, and
  Chalkidis}]{niklaus2023lextreme}
Joel Niklaus, Veton Matoshi, Pooja Rani, Andrea Galassi, Matthias Stürmer, and
  Ilias Chalkidis. 2023.
\newblock \href {http://arxiv.org/abs/2301.13126} {Lextreme: A multi-lingual
  and multi-task benchmark for the legal domain}.

\bibitem[{Ouyang et~al.(2022)Ouyang, Wu, Jiang, Almeida, Wainwright, Mishkin,
  Zhang, Agarwal, Slama, Ray, Schulman, Hilton, Kelton, Miller, Simens, Askell,
  Welinder, Christiano, Leike, and Lowe}]{instructgpt}
Long Ouyang, Jeff Wu, Xu~Jiang, Diogo Almeida, Carroll~L. Wainwright, Pamela
  Mishkin, Chong Zhang, Sandhini Agarwal, Katarina Slama, Alex Ray, John
  Schulman, Jacob Hilton, Fraser Kelton, Luke Miller, Maddie Simens, Amanda
  Askell, Peter Welinder, Paul Christiano, Jan Leike, and Ryan Lowe. 2022.
\newblock \href {https://doi.org/10.48550/ARXIV.2203.02155} {Training language
  models to follow instructions with human feedback}.

\bibitem[{Petroni et~al.(2019)Petroni, Rockt{\"a}schel, Riedel, Lewis, Bakhtin,
  Wu, and Miller}]{petroni-etal-2019-language}
Fabio Petroni, Tim Rockt{\"a}schel, Sebastian Riedel, Patrick Lewis, Anton
  Bakhtin, Yuxiang Wu, and Alexander Miller. 2019.
\newblock \href {https://doi.org/10.18653/v1/D19-1250} {Language models as
  knowledge bases?}
\newblock In \emph{Proceedings of the 2019 Conference on Empirical Methods in
  Natural Language Processing and the 9th International Joint Conference on
  Natural Language Processing (EMNLP-IJCNLP)}, pages 2463--2473, Hong Kong,
  China. Association for Computational Linguistics.

\bibitem[{Pfeiffer et~al.(2021)Pfeiffer, Vuli{\'c}, Gurevych, and
  Ruder}]{pfeiffer-etal-2021-unks}
Jonas Pfeiffer, Ivan Vuli{\'c}, Iryna Gurevych, and Sebastian Ruder. 2021.
\newblock \href {https://doi.org/10.18653/v1/2021.emnlp-main.800} {{UNK}s
  everywhere: {A}dapting multilingual language models to new scripts}.
\newblock In \emph{Proceedings of the 2021 Conference on Empirical Methods in
  Natural Language Processing}, pages 10186--10203, Online and Punta Cana,
  Dominican Republic. Association for Computational Linguistics.

\bibitem[{Radford and Narasimhan(2018)}]{Radford2018ImprovingLU}
Alec Radford and Karthik Narasimhan. 2018.
\newblock \href
  {https://s3-us-west-2.amazonaws.com/openai-assets/research-covers/language-unsupervised/language_understanding_paper.pdf}
  {Improving language understanding by generative pre-training}.

\bibitem[{Raffel et~al.(2020)Raffel, Shazeer, Roberts, Lee, Narang, Matena,
  Zhou, Li, and Liu}]{raffel2020}
Colin Raffel, Noam Shazeer, Adam Roberts, Katherine Lee, Sharan Narang, Michael
  Matena, Yanqi Zhou, Wei Li, and Peter~J. Liu. 2020.
\newblock \href {http://jmlr.org/papers/v21/20-074.html} {Exploring the limits
  of transfer learning with a unified text-to-text transformer}.
\newblock \emph{Journal of Machine Learning Research}, 21(140):1--67.

\bibitem[{R{\"u}ckl{\'e} et~al.(2021)R{\"u}ckl{\'e}, Geigle, Glockner, Beck,
  Pfeiffer, Reimers, and Gurevych}]{ruckle-etal-2021-adapterdrop}
Andreas R{\"u}ckl{\'e}, Gregor Geigle, Max Glockner, Tilman Beck, Jonas
  Pfeiffer, Nils Reimers, and Iryna Gurevych. 2021.
\newblock \href {https://doi.org/10.18653/v1/2021.emnlp-main.626}
  {{AdapterDrop}: {O}n the efficiency of adapters in transformers}.
\newblock In \emph{Proceedings of the 2021 Conference on Empirical Methods in
  Natural Language Processing}, pages 7930--7946, Online and Punta Cana,
  Dominican Republic. Association for Computational Linguistics.

\bibitem[{Sahlgren and Carlsson(2021)}]{sahlgren-2021}
Magnus Sahlgren and Fredrik Carlsson. 2021.
\newblock \href {https://doi.org/10.3389/frai.2021.682578} {The singleton
  fallacy: Why current critiques of language models miss the point}.
\newblock \emph{Frontiers in Artificial Intelligence}, 4.

\bibitem[{Stiennon et~al.(2020)Stiennon, Ouyang, Wu, Ziegler, Lowe, Voss,
  Radford, Amodei, and Christiano}]{Stiennon2020}
Nisan Stiennon, Long Ouyang, Jeffrey Wu, Daniel Ziegler, Ryan Lowe, Chelsea
  Voss, Alec Radford, Dario Amodei, and Paul~F Christiano. 2020.
\newblock \href
  {https://proceedings.neurips.cc/paper/2020/file/1f89885d556929e98d3ef9b86448f951-Paper.pdf}
  {Learning to summarize with human feedback}.
\newblock In \emph{Advances in Neural Information Processing Systems},
  volume~33, pages 3008--3021. Curran Associates, Inc.

\bibitem[{Tsarapatsanis and Aletras(2021)}]{tsarapatsanis-aletras-2021-ethical}
Dimitrios Tsarapatsanis and Nikolaos Aletras. 2021.
\newblock \href {https://doi.org/10.18653/v1/2021.findings-acl.314} {On the
  ethical limits of natural language processing on legal text}.
\newblock In \emph{Findings of the Association for Computational Linguistics:
  ACL-IJCNLP 2021}, pages 3590--3599, Online. Association for Computational
  Linguistics.

\bibitem[{Tuggener et~al.(2020)Tuggener, von D{\"a}niken, Peetz, and
  Cieliebak}]{tuggener-etal-2020-ledgar}
Don Tuggener, Pius von D{\"a}niken, Thomas Peetz, and Mark Cieliebak. 2020.
\newblock \href {https://aclanthology.org/2020.lrec-1.155} {{LEDGAR}: A
  large-scale multi-label corpus for text classification of legal provisions in
  contracts}.
\newblock In \emph{Proceedings of the Twelfth Language Resources and Evaluation
  Conference}, pages 1235--1241, Marseille, France. European Language Resources
  Association.

\bibitem[{Voorhees and Tice(2000)}]{207141}
Ellen Voorhees and D~Tice. 2000.
\newblock \href {https://tsapps.nist.gov/publication/get_pdf.cfm?pub_id=151446}
  {The trec-8 question answering track evaluation}.
\newblock 3. The TREC-8 Question Answering Track Evaluation.

\bibitem[{Wei et~al.(2021)Wei, Bosma, Zhao, Guu, Yu, Lester, Du, Dai, and
  Le}]{wei-etal-2022}
Jason Wei, Maarten Bosma, Vincent~Y. Zhao, Kelvin Guu, Adams~Wei Yu, Brian
  Lester, Nan Du, Andrew~M. Dai, and Quoc~V. Le. 2021.
\newblock \href {http://arxiv.org/abs/2109.01652} {Finetuned language models
  are zero-shot learners}.
\newblock \emph{CoRR}, abs/2109.01652.

\bibitem[{Wettig et~al.(2023)Wettig, Gao, Zhong, and Chen}]{wettig2022should}
Alexander Wettig, Tianyu Gao, Zexuan Zhong, and Danqi Chen. 2023.
\newblock \href {https://aclanthology.org/2023.eacl-main.217} {Should you mask
  15{\%} in masked language modeling?}
\newblock In \emph{Proceedings of the 17th Conference of the European Chapter
  of the Association for Computational Linguistics}, pages 2985--3000,
  Dubrovnik, Croatia. Association for Computational Linguistics.

\bibitem[{Xiao et~al.(2021)Xiao, Hu, Liu, Tu, and Sun}]{xiao-etal-2021}
Chaojun Xiao, Xueyu Hu, Zhiyuan Liu, Cunchao Tu, and Maosong Sun. 2021.
\newblock \href {http://arxiv.org/abs/2105.03887} {Lawformer: {A} pre-trained
  language model for chinese legal long documents}.
\newblock \emph{CoRR}, abs/2105.03887.

\bibitem[{Zheng et~al.(2021)Zheng, Guha, Anderson, Henderson, and
  Ho}]{zhengguha2021}
Lucia Zheng, Neel Guha, Brandon~R. Anderson, Peter Henderson, and Daniel~E. Ho.
  2021.
\newblock \href {http://arxiv.org/abs/2104.08671} {When does pretraining help?
  assessing self-supervised learning for law and the casehold dataset}.
\newblock In \emph{Proceedings of the 18th International Conference on
  Artificial Intelligence and Law}. Association for Computing Machinery.

\end{thebibliography}
\bibliographystyle{acl_natbib}

\appendix

\section{LegalLAMA Discussion}
\label{sec:legalama_discuss}

The \textsc{LegalLAMA} tasks cannot be resolved by laypersons or even law professionals that are not experts in the specific fields of law in many cases. Another consideration that often goes unspecified is that expertise is legal system-specific (e.g. US law differs widely from EU law), as do the distinctions between the academic and the practical knowledge of law (including potential sub-distinctions between different types of legal practitioners, e.g. litigation experts, contract drafting experts, due dilligence experts, etc.). Lastly, it is also important to note that legal systems can be clustered according to similarities or differences. Specifically:

\begin{itemize}[itemsep=-0.2em]
    \item For task \textbf{`ECHR Articles'}, both laypersons and lawyers who are not experts in human rights law (particularly ECHR) would perform at random chance level, since they lack knowledge of the ECHR in an article level. Providing the titles of the articles (Table~\ref{tab:echr_articles}), we can expect improved performance in case of rich context. Generally, the same can be said for the related task \textbf{`Legal Terminology (CoE)'}. Legal terminology is very particular to individual legal systems, and predicting the place of legal concepts within the ECHR would require a very high level of specialization. 
    \item For task \textbf{`Contractual Section Titles (US)'}, structural knowledge of US contracts would be necessary for the performance of this task with a high degree of accuracy. This is due to the fact that contracts often have some structural similarities, but also particular characteristics depending on the type of contract (e.g. employment, sale, credit). Laypersons would perform this task at random chance level. Practicing lawyers with contract drafting expertise would potentially have the highest performance in this task. Non-US lawyers with no contract drafting expertise would perform slightly higher than random chance level. The same considerations apply to the task \textbf{`Contract Types (US)'}.
    \item For tasks \textbf{`Crime Charges (US)'} and \textbf{`Criminal Code Sections (Canada)'}, both laypersons and lawyers who are not experts in criminal law (particularly US law and Canadian law) would perform at random chance level, since the legal concepts are very specific (e.g. manslaughter). Improved performance could be seen in cases where the masked terms are specifically defined.
    \item For tasks \textbf{`Legal Terminology (US)'} and \textbf{`Legal Terminology (EU)'}, the same discussion as above is applicable. Legal terminology is system-specific. There may be similar terms, but in the absence of knowledge relating to how such similarities may be interpreted, a non-expert lawyer would not perform such a task with a very high accuracy level.  
\end{itemize}

\subsection{ECtHR Articles}

We hereby provide details on the 13 ECtHR articles;

\begin{table}[h]
    \centering
    \resizebox{\columnwidth}{!}{
    \begin{tabular}{l|l}
    \bf ECHR Article & \bf Description (Title) \\
    \midrule
    Article 2 & Right to life \\
    Article 3 &  Prohibition of torture \\
    Article 5 &  Right to liberty and security \\
    Article 6 &  Right to a fair trial \\
    Article 7 &  No punishment without law \\
    Article 8 &  Right to respect for private and family life \\
    Article 9 &  Freedom of thought, conscience and religion \\
    Article 10 &  Freedom of expression \\
    Article 11 &  Freedom of assembly and association \\
    Article 13 &  Right to an effective remedy \\
    Article 14 &  Prohibition of discrimination \\
    Article 34 &  Individual applications \\
    Article 35 &  Admissibility criteria \\
    \end{tabular}
    }
    \caption{ECHR Articles}
    \label{tab:echr_articles}
\end{table}

\begin{table*}[t]
    \centering
    \resizebox{\textwidth}{!}{
    \setlength{\tabcolsep}{2.65pt}
    \begin{tabular}{l cc| cc| cc| cc| cc| cc| cc}
 & \multicolumn{2}{c}{RoBERTa-B} & \multicolumn{2}{c}{RoBERTa-L} & \multicolumn{2}{c}{LegalBERT} & \multicolumn{2}{c}{CL-BERT} & \multicolumn{2}{c}{PoL-BERT} & \multicolumn{2}{c}{LexLM-B} & \multicolumn{2}{c}{LexLM-L}\\
 \cmidrule(lr){2-15}
\textbf{Task} & P@1 & MRR & P@1 & MRR & P@1 & MRR & P@1 & MRR & P@1 & MRR & P@1 & MRR & P@1 & MRR\\
\cmidrule(lr){2-3} \cmidrule(lr){4-5} \cmidrule(lr){6-7} \cmidrule(lr){8-9} \cmidrule(lr){10-11} \cmidrule(lr){12-13} \cmidrule(lr){14-15} 

\textbf{ECHR Articles} & \cellcolor{gray!20}0.26 & 0.40 & \cellcolor{gray!20}0.27 & 0.41 & \cellcolor{gray!20}0.86 & 0.91 & \cellcolor{gray!20}0.23 & 0.38 & \cellcolor{gray!20}0.20 & 0.35 & \cellcolor{gray!20}0.86 & 0.91 & \cellcolor{gray!20}\textbf{0.91} & \textbf{0.94} \\ 
\textbf{Contract Sections} & \cellcolor{gray!20}0.20 & 0.40 & \cellcolor{gray!20}0.53 & 0.66 & \cellcolor{gray!20}0.77 & 0.85 & \cellcolor{gray!20}0.24 & 0.40 & \cellcolor{gray!20}0.51 & 0.65 & \cellcolor{gray!20}\textbf{0.78} & \textbf{0.86} & \cellcolor{gray!20}\textbf{0.78} & \textbf{0.86} \\
\textbf{Contract Types} & \cellcolor{gray!20}0.32 & 0.48 & \cellcolor{gray!20}0.34 & 0.50 & \cellcolor{gray!20}0.80 & 0.87 & \cellcolor{gray!20}0.42 & 0.55 & \cellcolor{gray!20}0.37 & 0.50 & \cellcolor{gray!20}0.82 & 0.89 & \cellcolor{gray!20}\textbf{0.85} & \textbf{0.91} \\
\textbf{Crime Charges (US)} & \cellcolor{gray!20}0.46 & 0.58 & \cellcolor{gray!20}0.54 & 0.65 & \cellcolor{gray!20}0.44 & 0.56 & \cellcolor{gray!20}0.51 & 0.63 & \cellcolor{gray!20}0.33 & 0.45 & \cellcolor{gray!20}0.56 & 0.67 & \cellcolor{gray!20}\textbf{0.61} & \textbf{0.71} \\
\textbf{Terminology (US)} & \cellcolor{gray!20}0.41 & 0.51 & \cellcolor{gray!20}0.49 & 0.58 & \cellcolor{gray!20}0.52 & 0.63 & \cellcolor{gray!20}0.58 & 0.69 & \cellcolor{gray!20}0.37 & 0.49 & \cellcolor{gray!20}0.64 & 0.74 & \cellcolor{gray!20}\textbf{0.70} & \textbf{0.79} \\
\textbf{Terminology (EU)} & \cellcolor{gray!20}0.34 & 0.47 & \cellcolor{gray!20}0.40 & 0.53 & \cellcolor{gray!20}0.51 & 0.64 & \cellcolor{gray!20}0.25 & 0.39 & \cellcolor{gray!20}0.25 & 0.38 & \cellcolor{gray!20}0.60 & 0.72 & \cellcolor{gray!20}\textbf{0.67} & \textbf{0.77} \\ 
\textbf{Terminology (CoE)} & \cellcolor{gray!20}0.43 & 0.54 & \cellcolor{gray!20}0.51 & 0.60 & \cellcolor{gray!20}0.69 & 0.78 & \cellcolor{gray!20}0.36 & 0.49 & \cellcolor{gray!20}0.30 & 0.41 & \cellcolor{gray!20}0.78 & 0.86 & \cellcolor{gray!20}\textbf{0.86} & \textbf{0.91} \\
\textbf{CC Sections} & \cellcolor{gray!20}0.36 & 0.45 & \cellcolor{gray!20}0.40 & 0.50 & \cellcolor{gray!20}0.53 & 0.59 & \cellcolor{gray!20}0.45 & 0.54 & \cellcolor{gray!20}0.46 & 0.53 & \cellcolor{gray!20}0.77 & 0.83 & \cellcolor{gray!20}\textbf{0.86} & \textbf{0.90} \\
\cmidrule(lr){2-15}
\textbf{Average} & \cellcolor{gray!20}0.33 & 0.47 & \cellcolor{gray!20}0.41 & 0.54 & \cellcolor{gray!20}0.61 & 0.71 & \cellcolor{gray!20}0.34 & 0.49 & \cellcolor{gray!20}0.32 & 0.46 & \cellcolor{gray!20}0.71 & 0.80 & \cellcolor{gray!20}\textbf{0.77} & \textbf{0.85}\\
\cmidrule(lr){2-15}
\textbf{Model Rank} & \multicolumn{2}{c|}{6} & \multicolumn{2}{c|}{4} & \multicolumn{2}{c|}{3} & \multicolumn{2}{c|}{5} & \multicolumn{2}{c|}{7} & \multicolumn{2}{c|}{2} & \multicolumn{2}{c}{1} \\

    \end{tabular}
    }
    \caption{P@1 and MRR results of the 7 examined PLMs on the 8 \textsc{LegalLAMA} tasks.}
    \label{tab:legal_lama_p1}
\end{table*}

\section{LexLM Pre-training Details}
\label{sec:lexlm_details}

For the newly released, LexLM models (LexLMs), we followed a series of best-practices in language model development literature:
\begin{enumerate}[label=(\alph*),itemsep=-0.3em]
    \item We warm-start (initialize) our models from the original RoBERTa checkpoints (base or large) of \citet{liu-2019-roberta}. Model recycling is a standard process followed by many~\cite{wei-etal-2022, instructgpt} to benefit from starting from an available ``well-trained'' PLM, instead from scratch (random).
    \item We train a new tokenizer of 50k BPEs based on the training subsets of \textsc{LeXFiles} to better cover legal language across all covered legal systems. Although, we reuse the original RoBERTa embeddings for all lexically overlapping tokens \cite{pfeiffer-etal-2021-unks}, i.e., we warm-start word embeddings for tokens that already exist in the original RoBERTa vocabulary, and use random ones for the rest.
    \item We continue pre-training our models on the diverse \textsc{LeXFiles} (Section~\ref{sec:corpus}) corpus for additional 1M steps with batches of 512 samples. We do initial warm-up steps for the first 5\% of the total training steps with a linearly increasing learning rate up to $1e\!-\!4$, and then follow a cosine decay scheduling, following recent trends. For half of the warm-up phase (2.5\%), the Transformer encoder is frozen, and only the embeddings, shared between input and output (MLM), are updated. We also use an increased 20/30\% masking rate, where also 100\% of the predictions are based on masked tokens, compared to \citet{devlin-etal-2019-bert}\footnote{\citeauthor{devlin-etal-2019-bert} --and many other follow-up work-- used a 15\% masking ratio, and a recipe of 80/10/10\% of predictions made across masked/randomly-replaced/original tokens.}  for base/large models respectively, based on the findings of~\citet{wettig2022should}. 
    \item For both training the tokenizer and the LexLM models, we use a sentence sampler with exponential smoothing of the sub-corpora sampling rate following \citet{conneau-etal-2019} and \citet{raffel2020}, since there is a disparate proportion of tokens across sub-corpora (Table~\ref{tab:pile}) and we aim to preserve per-corpus capacity, i.e., avoid overfitting to the majority (approx. 94\% of the total number or tokens) US-origin texts.
    \item We consider mixed cased models, similar to all recently developed large PLMs~\cite{liu-2019-roberta, raffel2020, brown-etal-gpt3}.
\end{enumerate}

We make LexLM models (base/large) publicly available alongside all intermediate checkpoints every 50k training steps on Hugging Face Hub.\footnote{\url{https://huggingface.co/lexlms}}

\section{Detailed Legal-LAMA results per tasks}
\label{app:detailed_lama}

Table~\ref{tab:legal_lama_p1} contains the same results as in Table~\ref{tab:legal_lama} with the addition of Precision@1 scores (P@1).
The reason why we decided to only present MRR results in the main paper is that the difference between MRR and P@1 does not change the ranking of the models, and P@1 does not account for minor variations in predictions.

For each task, we display detailed results per predicted terms for each model.
Table~\ref{tab:lama_echr} contains results on the 13 article numbers from the ECHR task.
Table~\ref{tab:lama_contract_sections} contains results on the 20 clause types from the Contract Section task.
Table~\ref{tab:lama_contract_types} contains results on the 16  types of contracts from the Contract Section task.
Table~\ref{tab:lama_us_crimes} contains results on the 11 topics from the Crime Charges (US) task. Each topic contains multiple labels.
Table~\ref{tab:lama_us_term} contains results on the 7 topics from the Terminology (US) task. Each topic contains multiple labels.
Table~\ref{tab:lama_eu_term} contains results on the 23 topics from the Terminology (EU) task. Each topic contains multiple labels.
Table~\ref{tab:lama_coe_term} contains results on the 12 articles from the Terminology (CoE) task. Each article contains multiple labels.
Table~\ref{tab:lama_ccc_term} contains results on the 43 sections from the Criminal Code Sections (Canada) task.

\begin{table*}[h!]
    \centering
    \resizebox{\textwidth}{!}{
    \setlength{\tabcolsep}{2.65pt}
    \begin{tabular}{l cc| cc| cc| cc| cc| cc| cc}
 & \multicolumn{2}{c}{RoBERTa-B} & \multicolumn{2}{c}{RoBERTa-L} & \multicolumn{2}{c}{LegalBERT} & \multicolumn{2}{c}{CL-BERT} & \multicolumn{2}{c}{PoL-BERT} & \multicolumn{2}{c}{LexLM-B} & \multicolumn{2}{c}{LexLM-L}\\
 \cmidrule(lr){2-15}
\textbf{ECHR Article} & P@1 & MRR & P@1 & MRR & P@1 & MRR & P@1 & MRR & P@1 & MRR & P@1 & MRR & P@1 & MRR\\
\cmidrule(lr){2-3} \cmidrule(lr){4-5} \cmidrule(lr){6-7} \cmidrule(lr){8-9} \cmidrule(lr){10-11} \cmidrule(lr){12-13} \cmidrule(lr){14-15} 

Art. 2 & \cellcolor{gray!20}0.87 & 0.91 & \cellcolor{gray!20}0.63 & 0.76 & \cellcolor{gray!20}0.87 & 0.92 & \cellcolor{gray!20}0.27 & 0.45 & \cellcolor{gray!20}0.29 & 0.51 & \cellcolor{gray!20}0.86 & 0.91 & \cellcolor{gray!20}0.91 & 0.94 \\ 
Art. 3 & \cellcolor{gray!20}0.23 & 0.56 & \cellcolor{gray!20}0.35 & 0.59 & \cellcolor{gray!20}0.93 & 0.96 & \cellcolor{gray!20}0.44 & 0.62 & \cellcolor{gray!20}0.32 & 0.54 & \cellcolor{gray!20}0.93 & 0.96 & \cellcolor{gray!20}0.96 & 0.97 \\
Art. 5 & \cellcolor{gray!20}0.35 & 0.56 & \cellcolor{gray!20}0.39 & 0.58 & \cellcolor{gray!20}0.83 & 0.89 & \cellcolor{gray!20}0.32 & 0.44 & \cellcolor{gray!20}0.20 & 0.41 & \cellcolor{gray!20}0.79 & 0.86 & \cellcolor{gray!20}0.88 & 0.92 \\ 
Art. 6 & \cellcolor{gray!20}0.27 & 0.40 & \cellcolor{gray!20}0.26 & 0.38 & \cellcolor{gray!20}0.93 & 0.96 & \cellcolor{gray!20}0.28 & 0.43 & \cellcolor{gray!20}0.18 & 0.36 & \cellcolor{gray!20}0.93 & 0.96 & \cellcolor{gray!20}0.94 & 0.96 \\ 
Art. 7 & \cellcolor{gray!20}0.15 & 0.38 & \cellcolor{gray!20}0.30 & 0.53 & \cellcolor{gray!20}0.53 & 0.72 & \cellcolor{gray!20}0.15 & 0.36 & \cellcolor{gray!20}0.29 & 0.49 & \cellcolor{gray!20}0.62 & 0.75 & \cellcolor{gray!20}0.74 & 0.83 \\ 
Art. 8 & \cellcolor{gray!20}0.16 & 0.28 & \cellcolor{gray!20}0.18 & 0.36 & \cellcolor{gray!20}0.89 & 0.93 & \cellcolor{gray!20}0.17 & 0.32 & \cellcolor{gray!20}0.13 & 0.30 & \cellcolor{gray!20}0.89 & 0.94 & \cellcolor{gray!20}0.91 & 0.95 \\ 
Art. 9 & \cellcolor{gray!20}0.33 & 0.46 & \cellcolor{gray!20}0.32 & 0.46 & \cellcolor{gray!20}0.83 & 0.89 & \cellcolor{gray!20}0.27 & 0.45 & \cellcolor{gray!20}0.27 & 0.45 & \cellcolor{gray!20}0.85 & 0.92 & \cellcolor{gray!20}0.95 & 0.97 \\
Art. 10 & \cellcolor{gray!20}0.23 & 0.34 & \cellcolor{gray!20}0.24 & 0.37 & \cellcolor{gray!20}0.84 & 0.90 & \cellcolor{gray!20}0.27 & 0.43 & \cellcolor{gray!20}0.21 & 0.33 & \cellcolor{gray!20}0.87 & 0.91 & \cellcolor{gray!20}0.90 & 0.93 \\ 
Art. 11 & \cellcolor{gray!20}0.25 & 0.33 & \cellcolor{gray!20}0.27 & 0.36 & \cellcolor{gray!20}0.94 & 0.96 & \cellcolor{gray!20}0.30 & 0.44 & \cellcolor{gray!20}0.23 & 0.34 & \cellcolor{gray!20}0.91 & 0.94 & \cellcolor{gray!20}0.97 & 0.99 \\ 
Art. 13 & \cellcolor{gray!20}0.28 & 0.36 & \cellcolor{gray!20}0.32 & 0.40 & \cellcolor{gray!20}0.89 & 0.94 & \cellcolor{gray!20}0.27 & 0.36 & \cellcolor{gray!20}0.26 & 0.39 & \cellcolor{gray!20}0.90 & 0.94 & \cellcolor{gray!20}0.92 & 0.95 \\ 
Art. 14 & \cellcolor{gray!20}0.14 & 0.24 & \cellcolor{gray!20}0.15 & 0.26 & \cellcolor{gray!20}0.85 & 0.91 & \cellcolor{gray!20}0.14 & 0.27 & \cellcolor{gray!20}0.07 & 0.19 & \cellcolor{gray!20}0.88 & 0.92 & \cellcolor{gray!20}0.90 & 0.94 \\  
Art. 34 & \cellcolor{gray!20}0.09 & 0.20 & \cellcolor{gray!20}0.08 & 0.19 & \cellcolor{gray!20}0.90 & 0.93 & \cellcolor{gray!20}0.08 & 0.17 & \cellcolor{gray!20}0.06 & 0.15 & \cellcolor{gray!20}0.90 & 0.94 & \cellcolor{gray!20}0.93 & 0.96 \\ 
Art. 35 & \cellcolor{gray!20}0.05 & 0.13 & \cellcolor{gray!20}0.06 & 0.17 & \cellcolor{gray!20}0.90 & 0.94 & \cellcolor{gray!20}0.05 & 0.13 & \cellcolor{gray!20}0.05 & 0.13 & \cellcolor{gray!20}0.88 & 0.93 & \cellcolor{gray!20}0.92 & 0.95 \\

\cmidrule(r){2-15}

\textbf{Average} & \cellcolor{gray!20}0.26 & 0.40 & \cellcolor{gray!20}0.27 & 0.41 & \cellcolor{gray!20}0.86 & 0.91 & \cellcolor{gray!20}0.23 & 0.38 & \cellcolor{gray!20}0.20 & 0.35 & \cellcolor{gray!20}0.86 & 0.91 & \cellcolor{gray!20}0.91 & 0.94 \\

    \end{tabular}
    }
    \caption{P@1 and MRR results of the 7 examined PLMs on the 13 article numbers from the ECHR task.}
    \label{tab:lama_echr}
\end{table*}

\begin{table*}[h!]
    \centering
    \resizebox{\textwidth}{!}{
    \setlength{\tabcolsep}{2.65pt}
    \begin{tabular}{l cc| cc| cc| cc| cc| cc| cc}
 & \multicolumn{2}{c}{RoBERTa-B} & \multicolumn{2}{c}{RoBERTa-L} & \multicolumn{2}{c}{LegalBERT} & \multicolumn{2}{c}{CL-BERT} & \multicolumn{2}{c}{PoL-BERT} & \multicolumn{2}{c}{LexLM-B} & \multicolumn{2}{c}{LexLM-L}\\
 \cmidrule(lr){2-15}
\textbf{Clause Type} & P@1 & MRR & P@1 & MRR & P@1 & MRR & P@1 & MRR & P@1 & MRR & P@1 & MRR & P@1 & MRR\\
\cmidrule(lr){2-3} \cmidrule(lr){4-5} \cmidrule(lr){6-7} \cmidrule(lr){8-9} \cmidrule(lr){10-11} \cmidrule(lr){12-13} \cmidrule(lr){14-15}

Arbitration & \cellcolor{gray!20}0.44 & 0.65 & \cellcolor{gray!20}0.97 & 0.98 & \cellcolor{gray!20}1.00 & 1.00 & \cellcolor{gray!20}0.83 & 0.91 & \cellcolor{gray!20}1.00 & 1.00 & \cellcolor{gray!20}1.00 & 1.00 & \cellcolor{gray!20}1.00 & 1.00 \\ 
Assignments & \cellcolor{gray!20}0.05 & 0.15 & \cellcolor{gray!20}0.34 & 0.49 & \cellcolor{gray!20}0.85 & 0.89 & \cellcolor{gray!20}0.01 & 0.12 & \cellcolor{gray!20}0.40 & 0.58 & \cellcolor{gray!20}0.90 & 0.94 & \cellcolor{gray!20}0.94 & 0.96 \\ 
Confidentiality & \cellcolor{gray!20}0.14 & 0.34 & \cellcolor{gray!20}0.73 & 0.84 & \cellcolor{gray!20}0.99 & 0.99 & \cellcolor{gray!20}0.14 & 0.34 & \cellcolor{gray!20}0.67 & 0.77 & \cellcolor{gray!20}0.99 & 0.99 & \cellcolor{gray!20}0.99 & 0.99 \\ 
Costs & \cellcolor{gray!20}0.00 & 0.22 & \cellcolor{gray!20}0.56 & 0.66 & \cellcolor{gray!20}0.78 & 0.89 & \cellcolor{gray!20}0.22 & 0.38 & \cellcolor{gray!20}0.33 & 0.54 & \cellcolor{gray!20}0.56 & 0.78 & \cellcolor{gray!20}0.67 & 0.80 \\ 
Definitions & \cellcolor{gray!20}1.00 & 1.00 & \cellcolor{gray!20}0.99 & 0.99 & \cellcolor{gray!20}0.78 & 0.84 & \cellcolor{gray!20}0.27 & 0.53 & \cellcolor{gray!20}0.75 & 0.85 & \cellcolor{gray!20}0.78 & 0.85 & \cellcolor{gray!20}0.81 & 0.87 \\ 
Disclosures & \cellcolor{gray!20}0.56 & 0.70 & \cellcolor{gray!20}0.37 & 0.50 & \cellcolor{gray!20}0.80 & 0.89 & \cellcolor{gray!20}0.02 & 0.16 & \cellcolor{gray!20}0.01 & 0.23 & \cellcolor{gray!20}0.65 & 0.80 & \cellcolor{gray!20}0.59 & 0.77 \\ 
Employment & \cellcolor{gray!20}0.42 & 0.69 & \cellcolor{gray!20}1.00 & 1.00 & \cellcolor{gray!20}0.92 & 0.96 & \cellcolor{gray!20}0.50 & 0.67 & \cellcolor{gray!20}0.65 & 0.80 & \cellcolor{gray!20}0.85 & 0.92 & \cellcolor{gray!20}1.00 & 1.00 \\ 
Enforceability & \cellcolor{gray!20}0.00 & 0.17 & \cellcolor{gray!20}0.26 & 0.37 & \cellcolor{gray!20}0.42 & 0.64 & \cellcolor{gray!20}0.00 & 0.06 & \cellcolor{gray!20}0.25 & 0.42 & \cellcolor{gray!20}0.33 & 0.54 & \cellcolor{gray!20}0.16 & 0.39 \\ 
Fees & \cellcolor{gray!20}0.12 & 0.50 & \cellcolor{gray!20}0.52 & 0.70 & \cellcolor{gray!20}0.43 & 0.62 & \cellcolor{gray!20}0.39 & 0.54 & \cellcolor{gray!20}0.38 & 0.60 & \cellcolor{gray!20}0.48 & 0.67 & \cellcolor{gray!20}0.51 & 0.69 \\ 
Indemnification & \cellcolor{gray!20}0.41 & 0.59 & \cellcolor{gray!20}0.70 & 0.80 & \cellcolor{gray!20}0.92 & 0.96 & \cellcolor{gray!20}0.10 & 0.34 & \cellcolor{gray!20}0.98 & 0.98 & \cellcolor{gray!20}0.96 & 0.98 & \cellcolor{gray!20}0.97 & 0.98 \\ 
Law & \cellcolor{gray!20}0.00 & 0.40 & \cellcolor{gray!20}0.21 & 0.57 & \cellcolor{gray!20}0.37 & 0.58 & \cellcolor{gray!20}0.87 & 0.92 & \cellcolor{gray!20}0.00 & 0.16 & \cellcolor{gray!20}0.79 & 0.87 & \cellcolor{gray!20}0.78 & 0.86 \\ 
Participations & \cellcolor{gray!20}0.04 & 0.20 & \cellcolor{gray!20}0.45 & 0.66 & \cellcolor{gray!20}0.82 & 0.90 & \cellcolor{gray!20}0.52 & 0.67 & \cellcolor{gray!20}0.38 & 0.59 & \cellcolor{gray!20}0.80 & 0.87 & \cellcolor{gray!20}0.82 & 0.89 \\ 
Remedies & \cellcolor{gray!20}0.05 & 0.25 & \cellcolor{gray!20}0.16 & 0.34 & \cellcolor{gray!20}0.92 & 0.96 & \cellcolor{gray!20}0.11 & 0.37 & \cellcolor{gray!20}0.52 & 0.71 & \cellcolor{gray!20}0.98 & 0.99 & \cellcolor{gray!20}0.99 & 0.99 \\ 
Representations & \cellcolor{gray!20}0.01 & 0.30 & \cellcolor{gray!20}0.43 & 0.62 & \cellcolor{gray!20}0.77 & 0.85 & \cellcolor{gray!20}0.17 & 0.46 & \cellcolor{gray!20}0.46 & 0.64 & \cellcolor{gray!20}0.86 & 0.91 & \cellcolor{gray!20}0.80 & 0.87 \\ 
Severability & \cellcolor{gray!20}0.02 & 0.17 & \cellcolor{gray!20}0.34 & 0.58 & \cellcolor{gray!20}0.99 & 0.99 & \cellcolor{gray!20}0.00 & 0.16 & \cellcolor{gray!20}0.97 & 0.98 & \cellcolor{gray!20}0.98 & 0.99 & \cellcolor{gray!20}0.98 & 0.99 \\ 
Solvency & \cellcolor{gray!20}0.09 & 0.22 & \cellcolor{gray!20}0.38 & 0.52 & \cellcolor{gray!20}0.94 & 0.97 & \cellcolor{gray!20}0.00 & 0.06 & \cellcolor{gray!20}0.11 & 0.26 & \cellcolor{gray!20}0.97 & 0.99 & \cellcolor{gray!20}0.97 & 0.99 \\ 
Taxes & \cellcolor{gray!20}0.29 & 0.59 & \cellcolor{gray!20}0.86 & 0.90 & \cellcolor{gray!20}0.99 & 0.99 & \cellcolor{gray!20}0.24 & 0.48 & \cellcolor{gray!20}0.56 & 0.68 & \cellcolor{gray!20}0.99 & 0.99 & \cellcolor{gray!20}0.99 & 0.99 \\ 
Termination & \cellcolor{gray!20}0.31 & 0.56 & \cellcolor{gray!20}0.60 & 0.77 & \cellcolor{gray!20}0.75 & 0.85 & \cellcolor{gray!20}0.22 & 0.45 & \cellcolor{gray!20}0.84 & 0.91 & \cellcolor{gray!20}0.80 & 0.89 & \cellcolor{gray!20}0.76 & 0.86 \\ 
Waivers & \cellcolor{gray!20}0.12 & 0.22 & \cellcolor{gray!20}0.59 & 0.67 & \cellcolor{gray!20}0.79 & 0.87 & \cellcolor{gray!20}0.00 & 0.07 & \cellcolor{gray!20}0.57 & 0.74 & \cellcolor{gray!20}0.94 & 0.95 & \cellcolor{gray!20}0.84 & 0.89 \\ 
Warranties & \cellcolor{gray!20}0.00 & 0.14 & \cellcolor{gray!20}0.05 & 0.26 & \cellcolor{gray!20}0.08 & 0.39 & \cellcolor{gray!20}0.14 & 0.33 & \cellcolor{gray!20}0.27 & 0.53 & \cellcolor{gray!20}0.05 & 0.36 & \cellcolor{gray!20}0.10 & 0.41 \\

\cmidrule(r){2-15}

\textbf{Average} & \cellcolor{gray!20}0.20 & 40.2 & \cellcolor{gray!20}0.53 & 0.66 & \cellcolor{gray!20}0.77 & 0.85 & \cellcolor{gray!20}0.24 & 0.40 & \cellcolor{gray!20}0.51 & 0.65 & \cellcolor{gray!20}0.78 & 0.86 & \cellcolor{gray!20}0.78 & 0.86 \\

    \end{tabular}
    }
    \caption{P@1 and MRR results of the 7 examined PLMs on the 20 clause types from the Contract Section task.}
    \label{tab:lama_contract_sections}
\end{table*}

\begin{table*}[h!]
    \centering
    \resizebox{\textwidth}{!}{
    \setlength{\tabcolsep}{2.65pt}
    \begin{tabular}{l cc| cc| cc| cc| cc| cc| cc}
 & \multicolumn{2}{c}{RoBERTa-B} & \multicolumn{2}{c}{RoBERTa-L} & \multicolumn{2}{c}{LegalBERT} & \multicolumn{2}{c}{CL-BERT} & \multicolumn{2}{c}{PoL-BERT} & \multicolumn{2}{c}{LexLM-B} & \multicolumn{2}{c}{LexLM-L}\\
 \cmidrule(lr){2-15}
\textbf{Contract Type} & P@1 & MRR & P@1 & MRR & P@1 & MRR & P@1 & MRR & P@1 & MRR & P@1 & MRR & P@1 & MRR\\
\cmidrule(lr){2-3} \cmidrule(lr){4-5} \cmidrule(lr){6-7} \cmidrule(lr){8-9} \cmidrule(lr){10-11} \cmidrule(lr){12-13} \cmidrule(lr){14-15}

Award & \cellcolor{gray!20}0.62 & 0.67 & \cellcolor{gray!20}0.62 & 0.70 & \cellcolor{gray!20}1.00 & 1.00 & \cellcolor{gray!20}0.54 & 0.60 & \cellcolor{gray!20}0.62 & 0.70 & \cellcolor{gray!20}1.00 & 1.00 & \cellcolor{gray!20}1.00 & 1.00 \\ 
Consulting & \cellcolor{gray!20}0.03 & 0.17 & \cellcolor{gray!20}0.10 & 0.23 & \cellcolor{gray!20}0.94 & 0.97 & \cellcolor{gray!20}0.08 & 0.29 & \cellcolor{gray!20}0.07 & 0.17 & \cellcolor{gray!20}0.81 & 0.87 & \cellcolor{gray!20}0.90 & 0.93 \\ 
Credit & \cellcolor{gray!20}0.57 & 0.72 & \cellcolor{gray!20}0.37 & 0.53 & \cellcolor{gray!20}0.97 & 0.98 & \cellcolor{gray!20}0.80 & 0.88 & \cellcolor{gray!20}0.55 & 0.77 & \cellcolor{gray!20}0.90 & 0.95 & \cellcolor{gray!20}0.95 & 0.98 \\ 
Employment & \cellcolor{gray!20}0.40 & 0.54 & \cellcolor{gray!20}0.30 & 0.44 & \cellcolor{gray!20}0.88 & 0.94 & \cellcolor{gray!20}0.63 & 0.73 & \cellcolor{gray!20}0.56 & 0.72 & \cellcolor{gray!20}0.99 & 0.99 & \cellcolor{gray!20}0.96 & 0.98 \\ 
Indemnity & \cellcolor{gray!20}0.08 & 0.34 & \cellcolor{gray!20}0.00 & 0.16 & \cellcolor{gray!20}0.62 & 0.71 & \cellcolor{gray!20}0.00 & 0.15 & \cellcolor{gray!20}0.00 & 0.11 & \cellcolor{gray!20}1.00 & 1.00 & \cellcolor{gray!20}1.00 & 1.00 \\ 
Letter & \cellcolor{gray!20}0.22 & 0.33 & \cellcolor{gray!20}0.24 & 0.34 & \cellcolor{gray!20}0.96 & 0.98 & \cellcolor{gray!20}0.76 & 0.87 & \cellcolor{gray!20}0.18 & 0.27 & \cellcolor{gray!20}0.77 & 0.88 & \cellcolor{gray!20}0.93 & 0.97 \\ 
License & \cellcolor{gray!20}0.40 & 0.62 & \cellcolor{gray!20}0.20 & 0.42 & \cellcolor{gray!20}0.63 & 0.76 & \cellcolor{gray!20}0.49 & 0.70 & \cellcolor{gray!20}0.31 & 0.44 & \cellcolor{gray!20}0.69 & 0.79 & \cellcolor{gray!20}0.86 & 0.91 \\ 
Loan & \cellcolor{gray!20}0.51 & 0.67 & \cellcolor{gray!20}0.72 & 0.84 & \cellcolor{gray!20}0.90 & 0.93 & \cellcolor{gray!20}0.72 & 0.83 & \cellcolor{gray!20}0.95 & 0.97 & \cellcolor{gray!20}0.90 & 0.94 & \cellcolor{gray!20}0.87 & 0.93 \\ 
Purchase & \cellcolor{gray!20}0.70 & 0.83 & \cellcolor{gray!20}0.59 & 0.68 & \cellcolor{gray!20}0.70 & 0.83 & \cellcolor{gray!20}0.52 & 0.68 & \cellcolor{gray!20}0.93 & 0.96 & \cellcolor{gray!20}0.89 & 0.92 & \cellcolor{gray!20}0.93 & 0.94 \\ 
Security & \cellcolor{gray!20}0.35 & 0.56 & \cellcolor{gray!20}0.70 & 0.80 & \cellcolor{gray!20}0.95 & 0.97 & \cellcolor{gray!20}0.59 & 0.75 & \cellcolor{gray!20}0.35 & 0.59 & \cellcolor{gray!20}0.97 & 0.99 & \cellcolor{gray!20}0.97 & 0.99 \\ 
Separation & \cellcolor{gray!20}0.12 & 0.26 & \cellcolor{gray!20}0.16 & 0.28 & \cellcolor{gray!20}0.66 & 0.77 & \cellcolor{gray!20}0.15 & 0.38 & \cellcolor{gray!20}0.07 & 0.21 & \cellcolor{gray!20}0.73 & 0.86 & \cellcolor{gray!20}0.71 & 0.82 \\ 
Services & \cellcolor{gray!20}0.24 & 0.45 & \cellcolor{gray!20}0.29 & 0.48 & \cellcolor{gray!20}0.52 & 0.67 & \cellcolor{gray!20}0.05 & 0.19 & \cellcolor{gray!20}0.38 & 0.54 & \cellcolor{gray!20}0.52 & 0.69 & \cellcolor{gray!20}0.52 & 0.69 \\ 
Settlement & \cellcolor{gray!20}0.49 & 0.63 & \cellcolor{gray!20}0.49 & 0.71 & \cellcolor{gray!20}0.70 & 0.80 & \cellcolor{gray!20}0.88 & 0.93 & \cellcolor{gray!20}0.58 & 0.72 & \cellcolor{gray!20}0.53 & 0.74 & \cellcolor{gray!20}0.65 & 0.80 \\ 
Supply & \cellcolor{gray!20}0.09 & 0.24 & \cellcolor{gray!20}0.35 & 0.51 & \cellcolor{gray!20}0.61 & 0.73 & \cellcolor{gray!20}0.09 & 0.19 & \cellcolor{gray!20}0.04 & 0.14 & \cellcolor{gray!20}0.70 & 0.77 & \cellcolor{gray!20}0.65 & 0.74 \\ 
Voting & \cellcolor{gray!20}0.00 & 0.13 & \cellcolor{gray!20}0.03 & 0.33 & \cellcolor{gray!20}1.00 & 1.00 & \cellcolor{gray!20}0.00 & 0.10 & \cellcolor{gray!20}0.00 & 0.13 & \cellcolor{gray!20}0.83 & 0.91 & \cellcolor{gray!20}0.90 & 0.95 \\

\cmidrule(r){2-15}

\textbf{Average} & \cellcolor{gray!20}0.32 & 0.48 & \cellcolor{gray!20}0.34 & 0.50 & \cellcolor{gray!20}0.80 & 0.87 & \cellcolor{gray!20}0.42 & 0.55 & \cellcolor{gray!20}0.37 & 0.50 & \cellcolor{gray!20}0.82 & 0.89 & \cellcolor{gray!20}0.85 & 0.91 \\

    \end{tabular}
    }
    \caption{P@1 and MRR results of the 7 examined PLMs on the 16 types of contracts from the Contract Types task.}
    \label{tab:lama_contract_types}
\end{table*}

\begin{table*}[h!]
    \centering
    \resizebox{\textwidth}{!}{
    \setlength{\tabcolsep}{2.65pt}
    \begin{tabular}{l cc| cc| cc| cc| cc| cc| cc}
 & \multicolumn{2}{c}{RoBERTa-B} & \multicolumn{2}{c}{RoBERTa-L} & \multicolumn{2}{c}{LegalBERT} & \multicolumn{2}{c}{CL-BERT} & \multicolumn{2}{c}{PoL-BERT} & \multicolumn{2}{c}{LexLM-B} & \multicolumn{2}{c}{LexLM-L}\\
 \cmidrule(lr){2-15}
\textbf{Crime Charges} & P@1 & MRR & P@1 & MRR & P@1 & MRR & P@1 & MRR & P@1 & MRR & P@1 & MRR & P@1 & MRR\\
\cmidrule(lr){2-3} \cmidrule(lr){4-5} \cmidrule(lr){6-7} \cmidrule(lr){8-9} \cmidrule(lr){10-11} \cmidrule(lr){12-13} \cmidrule(lr){14-15}

Children & \cellcolor{gray!20}0.69 & 0.78 & \cellcolor{gray!20}0.73 & 0.82 & \cellcolor{gray!20}0.47 & 0.61 & \cellcolor{gray!20}0.67 & 0.78 & \cellcolor{gray!20}0.45 & 0.60 & \cellcolor{gray!20}0.73 & 0.82 & \cellcolor{gray!20}0.77 & 0.85 \\ 
Computer & \cellcolor{gray!20}0.36 & 0.51 & \cellcolor{gray!20}0.46 & 0.62 & \cellcolor{gray!20}0.32 & 0.41 & \cellcolor{gray!20}0.42 & 0.53 & \cellcolor{gray!20}0.29 & 0.40 & \cellcolor{gray!20}0.44 & 0.56 & \cellcolor{gray!20}0.51 & 0.64 \\ 
Court-related & \cellcolor{gray!20}0.55 & 0.66 & \cellcolor{gray!20}0.57 & 0.69 & \cellcolor{gray!20}0.53 & 0.65 & \cellcolor{gray!20}0.61 & 0.73 & \cellcolor{gray!20}0.44 & 0.58 & \cellcolor{gray!20}0.63 & 0.74 & \cellcolor{gray!20}0.67 & 0.78 \\ 
Drug-related & \cellcolor{gray!20}0.40 & 0.53 & \cellcolor{gray!20}0.48 & 0.60 & \cellcolor{gray!20}0.31 & 0.44 & \cellcolor{gray!20}0.35 & 0.50 & \cellcolor{gray!20}0.26 & 0.38 & \cellcolor{gray!20}0.42 & 0.55 & \cellcolor{gray!20}0.46 & 0.60 \\ 
Wrongful Life Taking & \cellcolor{gray!20}0.50 & 0.64 & \cellcolor{gray!20}0.59 & 0.72 & \cellcolor{gray!20}0.59 & 0.71 & \cellcolor{gray!20}0.58 & 0.72 & \cellcolor{gray!20}0.31 & 0.47 & \cellcolor{gray!20}0.61 & 0.74 & \cellcolor{gray!20}0.63 & 0.76 \\ 
Mens Rea & \cellcolor{gray!20}0.56 & 0.64 & \cellcolor{gray!20}0.62 & 0.69 & \cellcolor{gray!20}0.55 & 0.65 & \cellcolor{gray!20}0.68 & 0.76 & \cellcolor{gray!20}0.47 & 0.59 & \cellcolor{gray!20}0.69 & 0.77 & \cellcolor{gray!20}0.75 & 0.82 \\ 
Monetary & \cellcolor{gray!20}0.40 & 0.51 & \cellcolor{gray!20}0.48 & 0.59 & \cellcolor{gray!20}0.52 & 0.63 & \cellcolor{gray!20}0.50 & 0.63 & \cellcolor{gray!20}0.30 & 0.44 & \cellcolor{gray!20}0.53 & 0.65 & \cellcolor{gray!20}0.61 & 0.72 \\ 
Pattern of Behavior & \cellcolor{gray!20}0.37 & 0.50 & \cellcolor{gray!20}0.48 & 0.59 & \cellcolor{gray!20}0.41 & 0.50 & \cellcolor{gray!20}0.44 & 0.57 & \cellcolor{gray!20}0.26 & 0.37 & \cellcolor{gray!20}0.52 & 0.62 & \cellcolor{gray!20}0.57 & 0.68 \\ 
Property & \cellcolor{gray!20}0.25 & 0.34 & \cellcolor{gray!20}0.36 & 0.43 & \cellcolor{gray!20}0.26 & 0.36 & \cellcolor{gray!20}0.32 & 0.41 & \cellcolor{gray!20}0.14 & 0.22 & \cellcolor{gray!20}0.40 & 0.46 & \cellcolor{gray!20}0.42 & 0.48 \\ 
Sex-related & \cellcolor{gray!20}0.55 & 0.65 & \cellcolor{gray!20}0.59 & 0.70 & \cellcolor{gray!20}0.47 & 0.59 & \cellcolor{gray!20}0.54 & 0.66 & \cellcolor{gray!20}0.36 & 0.48 & \cellcolor{gray!20}0.60 & 0.70 & \cellcolor{gray!20}0.66 & 0.75 \\ 
Violent & \cellcolor{gray!20}0.46 & 0.61 & \cellcolor{gray!20}0.57 & 0.70 & \cellcolor{gray!20}0.45 & 0.59 & \cellcolor{gray!20}0.54 & 0.69 & \cellcolor{gray!20}0.29 & 0.45 & \cellcolor{gray!20}0.58 & 0.72 & \cellcolor{gray!20}0.65 & 0.77 \\ 

\cmidrule(r){2-15}

\textbf{Average} & \cellcolor{gray!20}0.46 & 0.58 & \cellcolor{gray!20}0.54 & 0.65 & \cellcolor{gray!20}0.44 & 0.56 & \cellcolor{gray!20}0.51 & 0.63 & \cellcolor{gray!20}0.33 & 0.45 & \cellcolor{gray!20}0.56 & 0.67 & \cellcolor{gray!20}0.61 & 0.71 \\

    \end{tabular}
    }
    \caption{Results on the `Crime Charges (US)' \textsc{LegalLAMA} tasks. Results are clustered in Crime Topics.}
    \label{tab:lama_us_crimes}
\end{table*}

\begin{table*}[h!]
    \centering
    \resizebox{\textwidth}{!}{
    \setlength{\tabcolsep}{2.65pt}
    \begin{tabular}{l cc| cc| cc| cc| cc| cc| cc}
 & \multicolumn{2}{c}{RoBERTa-B} & \multicolumn{2}{c}{RoBERTa-L} & \multicolumn{2}{c}{LegalBERT} & \multicolumn{2}{c}{CL-BERT} & \multicolumn{2}{c}{PoL-BERT} & \multicolumn{2}{c}{LexLM-B} & \multicolumn{2}{c}{LexLM-L}\\
 \cmidrule(lr){2-15}
\textbf{Topic} & P@1 & MRR & P@1 & MRR & P@1 & MRR & P@1 & MRR & P@1 & MRR & P@1 & MRR & P@1 & MRR\\
\cmidrule(lr){2-3} \cmidrule(lr){4-5} \cmidrule(lr){6-7} \cmidrule(lr){8-9} \cmidrule(lr){10-11} \cmidrule(lr){12-13} \cmidrule(lr){14-15}

Business law & \cellcolor{gray!20}0.29 & 0.38 & \cellcolor{gray!20}0.37 & 0.45 & \cellcolor{gray!20}0.48 & 0.59 & \cellcolor{gray!20}0.59 & 0.70 & \cellcolor{gray!20}0.35 & 0.46 & \cellcolor{gray!20}0.59 & 0.71 & \cellcolor{gray!20}0.69 & 0.79 \\ 
Criminal law & \cellcolor{gray!20}0.39 & 0.49 & \cellcolor{gray!20}0.46 & 0.54 & \cellcolor{gray!20}0.48 & 0.58 & \cellcolor{gray!20}0.54 & 0.65 & \cellcolor{gray!20}0.32 & 0.45 & \cellcolor{gray!20}0.64 & 0.73 & \cellcolor{gray!20}0.67 & 0.76 \\ 
Employment law & \cellcolor{gray!20}0.47 & 0.60 & \cellcolor{gray!20}0.58 & 0.68 & \cellcolor{gray!20}0.47 & 0.60 & \cellcolor{gray!20}0.54 & 0.67 & \cellcolor{gray!20}0.41 & 0.54 & \cellcolor{gray!20}0.55 & 0.67 & \cellcolor{gray!20}0.65 & 0.76 \\ 
Family law & \cellcolor{gray!20}0.52 & 0.61 & \cellcolor{gray!20}0.59 & 0.67 & \cellcolor{gray!20}0.49 & 0.62 & \cellcolor{gray!20}0.66 & 0.77 & \cellcolor{gray!20}0.40 & 0.52 & \cellcolor{gray!20}0.75 & 0.84 & \cellcolor{gray!20}0.82 & 0.88 \\ 
Immigration & \cellcolor{gray!20}0.48 & 0.57 & \cellcolor{gray!20}0.54 & 0.62 & \cellcolor{gray!20}0.58 & 0.67 & \cellcolor{gray!20}0.55 & 0.65 & \cellcolor{gray!20}0.38 & 0.48 & \cellcolor{gray!20}0.65 & 0.74 & \cellcolor{gray!20}0.72 & 0.80 \\ 
Landlord-tenant law & \cellcolor{gray!20}0.37 & 0.46 & \cellcolor{gray!20}0.44 & 0.52 & \cellcolor{gray!20}0.64 & 0.73 & \cellcolor{gray!20}0.69 & 0.77 & \cellcolor{gray!20}0.42 & 0.52 & \cellcolor{gray!20}0.75 & 0.82 & \cellcolor{gray!20}0.80 & 0.86 \\ 
Bankruptcy & \cellcolor{gray!20}0.37 & 0.49 & \cellcolor{gray!20}0.43 & 0.55 & \cellcolor{gray!20}0.48 & 0.59 & \cellcolor{gray!20}0.49 & 0.62 & \cellcolor{gray!20}0.34 & 0.47 & \cellcolor{gray!20}0.53 & 0.66 & \cellcolor{gray!20}0.59 & 0.71 \\

\cmidrule(r){2-15}

\textbf{Average} & \cellcolor{gray!20}0.41 & 0.51 & \cellcolor{gray!20}0.49 & 0.58 & \cellcolor{gray!20}0.52 & 0.63 & \cellcolor{gray!20}0.58 & 0.69 & \cellcolor{gray!20}0.37 & 0.49 & \cellcolor{gray!20}0.64 & 0.74 & \cellcolor{gray!20}0.70 & 0.79 \\

    \end{tabular}
    }
    \caption{Results on the `Terminology (US)' \textsc{LegalLAMA} task. Results are clustered in Law Topics.}
    \label{tab:lama_us_term}
\end{table*}

\begin{table*}[h!]
    \centering
    \resizebox{\textwidth}{!}{
    \setlength{\tabcolsep}{2.65pt}
    \begin{tabular}{l cc| cc| cc| cc| cc| cc| cc}
 & \multicolumn{2}{c}{RoBERTa-B} & \multicolumn{2}{c}{RoBERTa-L} & \multicolumn{2}{c}{LegalBERT} & \multicolumn{2}{c}{CL-BERT} & \multicolumn{2}{c}{PoL-BERT} & \multicolumn{2}{c}{LexLM-B} & \multicolumn{2}{c}{LexLM-L}\\
 \cmidrule(lr){2-15}
\textbf{Topic} & P@1 & MRR & P@1 & MRR & P@1 & MRR & P@1 & MRR & P@1 & MRR & P@1 & MRR & P@1 & MRR\\
\cmidrule(lr){2-3} \cmidrule(lr){4-5} \cmidrule(lr){6-7} \cmidrule(lr){8-9} \cmidrule(lr){10-11} \cmidrule(lr){12-13} \cmidrule(lr){14-15}

Accession & \cellcolor{gray!20}0.32 & 0.45 & \cellcolor{gray!20}0.57 & 0.68 & \cellcolor{gray!20}0.93 & 0.95 & \cellcolor{gray!20}0.46 & 0.55 & \cellcolor{gray!20}0.87 & 0.90 & \cellcolor{gray!20}0.80 & 0.88 & \cellcolor{gray!20}0.80 & 0.89 \\ 
Administrative cooperation & \cellcolor{gray!20}0.15 & 0.33 & \cellcolor{gray!20}0.23 & 0.40 & \cellcolor{gray!20}0.53 & 0.69 & \cellcolor{gray!20}0.12 & 0.27 & \cellcolor{gray!20}0.19 & 0.32 & \cellcolor{gray!20}0.65 & 0.79 & \cellcolor{gray!20}0.82 & 0.89 \\ 
Approximation of laws & \cellcolor{gray!20}0.46 & 0.54 & \cellcolor{gray!20}0.54 & 0.58 & \cellcolor{gray!20}0.36 & 0.47 & \cellcolor{gray!20}0.18 & 0.32 & \cellcolor{gray!20}0.08 & 0.23 & \cellcolor{gray!20}0.67 & 0.73 & \cellcolor{gray!20}0.72 & 0.79 \\ 
Area of freedom, security and justice & \cellcolor{gray!20}0.14 & 0.27 & \cellcolor{gray!20}0.13 & 0.28 & \cellcolor{gray!20}0.11 & 0.24 & \cellcolor{gray!20}0.14 & 0.28 & \cellcolor{gray!20}0.11 & 0.25 & \cellcolor{gray!20}0.13 & 0.27 & \cellcolor{gray!20}0.19 & 0.34 \\ 
Citizenship of the union & \cellcolor{gray!20}0.40 & 0.60 & \cellcolor{gray!20}0.47 & 0.64 & \cellcolor{gray!20}0.26 & 0.45 & \cellcolor{gray!20}0.12 & 0.30 & \cellcolor{gray!20}0.31 & 0.47 & \cellcolor{gray!20}0.50 & 0.70 & \cellcolor{gray!20}0.53 & 0.72 \\ 
Competition & \cellcolor{gray!20}0.50 & 0.68 & \cellcolor{gray!20}0.75 & 0.80 & \cellcolor{gray!20}0.84 & 0.90 & \cellcolor{gray!20}0.52 & 0.62 & \cellcolor{gray!20}0.52 & 0.62 & \cellcolor{gray!20}0.88 & 0.89 & \cellcolor{gray!20}0.88 & 0.89 \\ 
Consumer protection & \cellcolor{gray!20}0.40 & 0.57 & \cellcolor{gray!20}0.50 & 0.62 & \cellcolor{gray!20}0.45 & 0.58 & \cellcolor{gray!20}0.28 & 0.42 & \cellcolor{gray!20}0.20 & 0.37 & \cellcolor{gray!20}0.25 & 0.42 & \cellcolor{gray!20}0.40 & 0.54 \\ 
Data protection & \cellcolor{gray!20}0.47 & 0.63 & \cellcolor{gray!20}0.61 & 0.73 & \cellcolor{gray!20}0.64 & 0.75 & \cellcolor{gray!20}0.17 & 0.28 & \cellcolor{gray!20}0.20 & 0.35 & \cellcolor{gray!20}0.66 & 0.76 & \cellcolor{gray!20}0.73 & 0.82 \\ 
External relations & \cellcolor{gray!20}0.30 & 0.45 & \cellcolor{gray!20}0.40 & 0.61 & \cellcolor{gray!20}0.38 & 0.55 & \cellcolor{gray!20}0.19 & 0.29 & \cellcolor{gray!20}0.09 & 0.22 & \cellcolor{gray!20}0.40 & 0.61 & \cellcolor{gray!20}0.55 & 0.68 \\ 
Free movement of capital & \cellcolor{gray!20}0.42 & 0.45 & \cellcolor{gray!20}0.42 & 0.45 & \cellcolor{gray!20}0.18 & 0.38 & \cellcolor{gray!20}0.11 & 0.26 & \cellcolor{gray!20}0.08 & 0.22 & \cellcolor{gray!20}0.33 & 0.53 & \cellcolor{gray!20}0.33 & 0.59 \\ 
Free movement of goods & \cellcolor{gray!20}0.25 & 0.37 & \cellcolor{gray!20}0.25 & 0.35 & \cellcolor{gray!20}0.32 & 0.48 & \cellcolor{gray!20}0.21 & 0.34 & \cellcolor{gray!20}0.18 & 0.31 & \cellcolor{gray!20}0.62 & 0.74 & \cellcolor{gray!20}0.38 & 0.58 \\ 
Freedom of establishment & \cellcolor{gray!20}0.22 & 0.34 & \cellcolor{gray!20}0.42 & 0.50 & \cellcolor{gray!20}0.64 & 0.75 & \cellcolor{gray!20}0.33 & 0.43 & \cellcolor{gray!20}0.29 & 0.40 & \cellcolor{gray!20}0.81 & 0.88 & \cellcolor{gray!20}0.94 & 0.95 \\ 
Freedom of movement for workers & \cellcolor{gray!20}0.22 & 0.34 & \cellcolor{gray!20}0.35 & 0.41 & \cellcolor{gray!20}0.19 & 0.35 & \cellcolor{gray!20}0.12 & 0.23 & \cellcolor{gray!20}0.11 & 0.22 & \cellcolor{gray!20}0.43 & 0.56 & \cellcolor{gray!20}0.38 & 0.55 \\ 
Freedom to provide services & \cellcolor{gray!20}0.07 & 0.20 & \cellcolor{gray!20}0.04 & 0.23 & \cellcolor{gray!20}0.23 & 0.40 & \cellcolor{gray!20}0.10 & 0.24 & \cellcolor{gray!20}0.15 & 0.29 & \cellcolor{gray!20}0.39 & 0.58 & \cellcolor{gray!20}0.54 & 0.67 \\ 
Fundamental rights & \cellcolor{gray!20}0.60 & 0.73 & \cellcolor{gray!20}0.69 & 0.81 & \cellcolor{gray!20}0.89 & 0.93 & \cellcolor{gray!20}0.26 & 0.37 & \cellcolor{gray!20}0.22 & 0.36 & \cellcolor{gray!20}0.84 & 0.90 & \cellcolor{gray!20}0.83 & 0.89 \\ 
Internal market & \cellcolor{gray!20}0.00 & 0.24 & \cellcolor{gray!20}0.20 & 0.40 & \cellcolor{gray!20}0.94 & 0.96 & \cellcolor{gray!20}0.26 & 0.36 & \cellcolor{gray!20}0.40 & 0.55 & \cellcolor{gray!20}0.40 & 0.62 & \cellcolor{gray!20}0.70 & 0.77 \\ 
Non-contractual liability & \cellcolor{gray!20}0.09 & 0.19 & \cellcolor{gray!20}0.09 & 0.20 & \cellcolor{gray!20}0.19 & 0.35 & \cellcolor{gray!20}0.19 & 0.40 & \cellcolor{gray!20}0.10 & 0.23 & \cellcolor{gray!20}0.30 & 0.49 & \cellcolor{gray!20}0.55 & 0.70 \\ 
Non-discrimination & \cellcolor{gray!20}0.00 & 0.24 & \cellcolor{gray!20}0.00 & 0.25 & \cellcolor{gray!20}0.50 & 0.68 & \cellcolor{gray!20}0.29 & 0.48 & \cellcolor{gray!20}0.10 & 0.26 & \cellcolor{gray!20}0.67 & 0.83 & \cellcolor{gray!20}0.33 & 0.67 \\ 
Privileges and immunities & \cellcolor{gray!20}0.17 & 0.27 & \cellcolor{gray!20}0.12 & 0.24 & \cellcolor{gray!20}0.63 & 0.77 & \cellcolor{gray!20}0.25 & 0.36 & \cellcolor{gray!20}0.20 & 0.35 & \cellcolor{gray!20}0.81 & 0.88 & \cellcolor{gray!20}0.81 & 0.87 \\ 
Procedural provisions & \cellcolor{gray!20}0.53 & 0.66 & \cellcolor{gray!20}0.63 & 0.75 & \cellcolor{gray!20}0.68 & 0.80 & \cellcolor{gray!20}0.61 & 0.73 & \cellcolor{gray!20}0.42 & 0.56 & \cellcolor{gray!20}0.71 & 0.82 & \cellcolor{gray!20}0.75 & 0.84 \\ 
Public health & \cellcolor{gray!20}0.62 & 0.80 & \cellcolor{gray!20}0.50 & 0.72 & \cellcolor{gray!20}0.68 & 0.79 & \cellcolor{gray!20}0.38 & 0.58 & \cellcolor{gray!20}0.28 & 0.48 & \cellcolor{gray!20}0.54 & 0.75 & \cellcolor{gray!20}0.92 & 0.96 \\ 
Safeguard measures & \cellcolor{gray!20}0.50 & 0.52 & \cellcolor{gray!20}0.50 & 0.58 & \cellcolor{gray!20}0.64 & 0.76 & \cellcolor{gray!20}0.31 & 0.39 & \cellcolor{gray!20}0.42 & 0.52 & \cellcolor{gray!20}0.75 & 0.88 & \cellcolor{gray!20}1.00 & 1.00 \\ 
Social policy & \cellcolor{gray!20}0.75 & 0.78 & \cellcolor{gray!20}0.75 & 0.81 & \cellcolor{gray!20}0.42 & 0.54 & \cellcolor{gray!20}0.22 & 0.37 & \cellcolor{gray!20}0.15 & 0.32 & \cellcolor{gray!20}0.75 & 0.83 & \cellcolor{gray!20}1.00 & 1.00 \\  

\cmidrule(r){2-15}

\textbf{Average} & \cellcolor{gray!20}0.34 & 0.47 & \cellcolor{gray!20}0.40 & 0.53 & \cellcolor{gray!20}0.51 & 0.64 & \cellcolor{gray!20}0.25 & 0.39 & \cellcolor{gray!20}0.25 & 0.38 & \cellcolor{gray!20}0.60 & 0.72 & \cellcolor{gray!20}0.67 & 0.77 \\

    \end{tabular}
    }
    \caption{Results on the `Terminology (EU)' \textsc{LegalLAMA} task. Results are clustered in Law Topics.}
    \label{tab:lama_eu_term}
\end{table*}

\begin{table*}[h!]
    \centering
    \resizebox{\textwidth}{!}{
    \setlength{\tabcolsep}{2.65pt}
    \begin{tabular}{l cc| cc| cc| cc| cc| cc| cc}
 & \multicolumn{2}{c}{RoBERTa-B} & \multicolumn{2}{c}{RoBERTa-L} & \multicolumn{2}{c}{LegalBERT} & \multicolumn{2}{c}{CL-BERT} & \multicolumn{2}{c}{PoL-BERT} & \multicolumn{2}{c}{LexLM-B} & \multicolumn{2}{c}{LexLM-L}\\
 \cmidrule(lr){2-15}
\textbf{Article} & P@1 & MRR & P@1 & MRR & P@1 & MRR & P@1 & MRR & P@1 & MRR & P@1 & MRR & P@1 & MRR\\
\cmidrule(lr){2-3} \cmidrule(lr){4-5} \cmidrule(lr){6-7} \cmidrule(lr){8-9} \cmidrule(lr){10-11} \cmidrule(lr){12-13} \cmidrule(lr){14-15}

 Art. 2 & \cellcolor{gray!20}0.46 & 0.57 & \cellcolor{gray!20}0.52 & 0.63 & \cellcolor{gray!20}0.72 & 0.82 & \cellcolor{gray!20}0.37 & 0.51 & \cellcolor{gray!20}0.36 & 0.47 & \cellcolor{gray!20}0.80 & 0.87 & \cellcolor{gray!20}0.90 & 0.94 \\ 
Art. 3 & \cellcolor{gray!20}0.51 & 0.61 & \cellcolor{gray!20}0.58 & 0.69 & \cellcolor{gray!20}0.80 & 0.87 & \cellcolor{gray!20}0.40 & 0.54 & \cellcolor{gray!20}0.34 & 0.45 & \cellcolor{gray!20}0.83 & 0.90 & \cellcolor{gray!20}0.89 & 0.93 \\  
Art. 5 & \cellcolor{gray!20}0.39 & 0.51 & \cellcolor{gray!20}0.46 & 0.57 & \cellcolor{gray!20}0.56 & 0.69 & \cellcolor{gray!20}0.36 & 0.48 & \cellcolor{gray!20}0.25 & 0.38 & \cellcolor{gray!20}0.63 & 0.75 & \cellcolor{gray!20}0.74 & 0.83 \\ 
Art. 6 & \cellcolor{gray!20}0.42 & 0.55 & \cellcolor{gray!20}0.49 & 0.62 & \cellcolor{gray!20}0.68 & 0.77 & \cellcolor{gray!20}0.43 & 0.55 & \cellcolor{gray!20}0.36 & 0.49 & \cellcolor{gray!20}0.77 & 0.85 & \cellcolor{gray!20}0.82 & 0.89 \\ 
Art. 7 & \cellcolor{gray!20}0.71 & 0.78 & \cellcolor{gray!20}0.82 & 0.86 & \cellcolor{gray!20}0.89 & 0.93 & \cellcolor{gray!20}0.36 & 0.59 & \cellcolor{gray!20}0.44 & 0.52 & \cellcolor{gray!20}0.88 & 0.93 & \cellcolor{gray!20}0.91 & 0.94 \\ 
Art. 8 & \cellcolor{gray!20}0.35 & 0.47 & \cellcolor{gray!20}0.45 & 0.56 & \cellcolor{gray!20}0.62 & 0.71 & \cellcolor{gray!20}0.29 & 0.41 & \cellcolor{gray!20}0.26 & 0.36 & \cellcolor{gray!20}0.73 & 0.82 & \cellcolor{gray!20}0.84 & 0.90 \\ 
Art. 9 & \cellcolor{gray!20}0.49 & 0.57 & \cellcolor{gray!20}0.56 & 0.64 & \cellcolor{gray!20}0.67 & 0.76 & \cellcolor{gray!20}0.43 & 0.53 & \cellcolor{gray!20}0.33 & 0.44 & \cellcolor{gray!20}0.79 & 0.86 & \cellcolor{gray!20}0.85 & 0.91 \\
Art. 10 & \cellcolor{gray!20}0.30 & 0.43 & \cellcolor{gray!20}0.41 & 0.52 & \cellcolor{gray!20}0.57 & 0.69 & \cellcolor{gray!20}0.25 & 0.37 & \cellcolor{gray!20}0.20 & 0.31 & \cellcolor{gray!20}0.73 & 0.82 & \cellcolor{gray!20}0.84 & 0.90 \\ 
Art. 11 & \cellcolor{gray!20}0.32 & 0.44 & \cellcolor{gray!20}0.42 & 0.52 & \cellcolor{gray!20}0.66 & 0.75 & \cellcolor{gray!20}0.29 & 0.40 & \cellcolor{gray!20}0.23 & 0.34 & \cellcolor{gray!20}0.74 & 0.84 & \cellcolor{gray!20}0.87 & 0.92 \\ 
Art. 13 & \cellcolor{gray!20}0.44 & 0.61 & \cellcolor{gray!20}0.55 & 0.69 & \cellcolor{gray!20}0.78 & 0.86 & \cellcolor{gray!20}0.38 & 0.56 & \cellcolor{gray!20}0.27 & 0.45 & \cellcolor{gray!20}0.86 & 0.90 & \cellcolor{gray!20}0.91 & 0.94 \\ 
Art. 14 & \cellcolor{gray!20}0.72 & 0.80 & \cellcolor{gray!20}0.79 & 0.85 & \cellcolor{gray!20}0.80 & 0.86 & \cellcolor{gray!20}0.69 & 0.78 & \cellcolor{gray!20}0.52 & 0.63 & \cellcolor{gray!20}0.84 & 0.89 & \cellcolor{gray!20}0.91 & 0.94 \\
Art. 35 & \cellcolor{gray!20}0.14 & 0.21 & \cellcolor{gray!20}0.18 & 0.24 & \cellcolor{gray!20}0.61 & 0.71 & \cellcolor{gray!20}0.14 & 0.26 & \cellcolor{gray!20}0.09 & 0.18 & \cellcolor{gray!20}0.78 & 0.85 & \cellcolor{gray!20}0.89 & 0.93 \\

\cmidrule(r){2-15}

\textbf{Average} & \cellcolor{gray!20}0.43 & 0.54 & \cellcolor{gray!20}0.51 & 0.61 & \cellcolor{gray!20}0.69 & 0.79 & \cellcolor{gray!20}0.36 & 0.49 & \cellcolor{gray!20}0.30 & 0.41 & \cellcolor{gray!20}0.78 & 0.86 & \cellcolor{gray!20}0.86 & 0.91 \\

    \end{tabular}
    }
    \caption{Results on the `Terminology (CoE)' \textsc{LegalLAMA} task. Results are clustered by Article.}
    \label{tab:lama_coe_term}
\end{table*}

\begin{table*}[h!]
    \centering
    \resizebox{\textwidth}{!}{
    \setlength{\tabcolsep}{2.65pt}
    \begin{tabular}{l cc| cc| cc| cc| cc| cc| cc}
 & \multicolumn{2}{c}{RoBERTa-B} & \multicolumn{2}{c}{RoBERTa-L} & \multicolumn{2}{c}{LegalBERT} & \multicolumn{2}{c}{CL-BERT} & \multicolumn{2}{c}{PoL-BERT} & \multicolumn{2}{c}{LexLM-B} & \multicolumn{2}{c}{LexLM-L}\\
 \cmidrule(lr){2-15}
\textbf{Section} & P@1 & MRR & P@1 & MRR & P@1 & MRR & P@1 & MRR & P@1 & MRR & P@1 & MRR & P@1 & MRR\\
\cmidrule(lr){2-3} \cmidrule(lr){4-5} \cmidrule(lr){6-7} \cmidrule(lr){8-9} \cmidrule(lr){10-11} \cmidrule(lr){12-13} \cmidrule(lr){14-15}

16 & \cellcolor{gray!20}0.00 & 0.08 & \cellcolor{gray!20}0.00 & 0.04 & \cellcolor{gray!20}0.00 & 0.04 & \cellcolor{gray!20}0.00 & 0.10 & \cellcolor{gray!20}0.00 & 0.08 & \cellcolor{gray!20}0.50 & 0.62 & \cellcolor{gray!20}1.00 & 1.00 \\
21 & \cellcolor{gray!20}0.23 & 0.41 & \cellcolor{gray!20}0.37 & 0.47 & \cellcolor{gray!20}0.46 & 0.56 & \cellcolor{gray!20}0.43 & 0.56 & \cellcolor{gray!20}0.44 & 0.55 & \cellcolor{gray!20}0.94 & 0.96 & \cellcolor{gray!20}0.97 & 0.99 \\
85 & \cellcolor{gray!20}0.46 & 0.51 & \cellcolor{gray!20}0.31 & 0.42 & \cellcolor{gray!20}0.30 & 0.41 & \cellcolor{gray!20}0.29 & 0.37 & \cellcolor{gray!20}0.30 & 0.39 & \cellcolor{gray!20}0.40 & 0.52 & \cellcolor{gray!20}0.57 & 0.69 \\ 
86 & \cellcolor{gray!20}0.38 & 0.53 & \cellcolor{gray!20}0.38 & 0.50 & \cellcolor{gray!20}0.50 & 0.62 & \cellcolor{gray!20}0.50 & 0.54 & \cellcolor{gray!20}0.50 & 0.53 & \cellcolor{gray!20}0.50 & 0.71 & \cellcolor{gray!20}0.50 & 0.66 \\ 
87  & \cellcolor{gray!20}0.75 & 0.78 & \cellcolor{gray!20}0.50 & 0.62 & \cellcolor{gray!20}0.75 & 0.79 & \cellcolor{gray!20}0.50 & 0.65 & \cellcolor{gray!20}0.75 & 0.82 & \cellcolor{gray!20}0.75 & 0.83 & \cellcolor{gray!20}0.75 & 0.80 \\ 
88.23 & \cellcolor{gray!20}0.25 & 0.34 & \cellcolor{gray!20}0.33 & 0.38 & \cellcolor{gray!20}0.33 & 0.38 & \cellcolor{gray!20}0.33 & 0.39 & \cellcolor{gray!20}0.33 & 0.42 & \cellcolor{gray!20}0.33 & 0.40 & \cellcolor{gray!20}0.33 & 0.38 \\ 
95 & \cellcolor{gray!20}0.48 & 0.54 & \cellcolor{gray!20}0.52 & 0.56 & \cellcolor{gray!20}0.52 & 0.55 & \cellcolor{gray!20}0.46 & 0.52 & \cellcolor{gray!20}0.45 & 0.49 & \cellcolor{gray!20}0.79 & 0.84 & \cellcolor{gray!20}0.80 & 0.85 \\
122 & \cellcolor{gray!20}0.17 & 0.19 & \cellcolor{gray!20}0.11 & 0.15 & \cellcolor{gray!20}0.17 & 0.18 & \cellcolor{gray!20}0.12 & 0.15 & \cellcolor{gray!20}0.12 & 0.16 & \cellcolor{gray!20}0.50 & 0.67 & \cellcolor{gray!20}0.83 & 0.86 \\ 
145 & \cellcolor{gray!20}0.25 & 0.38 & \cellcolor{gray!20}0.25 & 0.40 & \cellcolor{gray!20}0.44 & 0.51 & \cellcolor{gray!20}0.38 & 0.50 & \cellcolor{gray!20}0.50 & 0.55 & \cellcolor{gray!20}0.62 & 0.71 & \cellcolor{gray!20}0.88 & 0.90 \\ 
151 & \cellcolor{gray!20}0.59 & 0.61 & \cellcolor{gray!20}0.89 & 0.91 & \cellcolor{gray!20}0.62 & 0.64 & \cellcolor{gray!20}0.04 & 0.34 & \cellcolor{gray!20}0.02 & 0.32 & \cellcolor{gray!20}0.91 & 0.92 & \cellcolor{gray!20}0.91 & 0.92 \\ 
152 & \cellcolor{gray!20}1.00 & 1.00 & \cellcolor{gray!20}1.00 & 1.00 & \cellcolor{gray!20}1.00 & 1.00 & \cellcolor{gray!20}1.00 & 1.00 & \cellcolor{gray!20}0.50 & 0.75 & \cellcolor{gray!20}1.00 & 1.00 & \cellcolor{gray!20}1.00 & 1.00 \\ 
163 & \cellcolor{gray!20}0.50 & 0.52 & \cellcolor{gray!20}0.50 & 0.51 & \cellcolor{gray!20}0.50 & 0.51 & \cellcolor{gray!20}0.50 & 0.51 & \cellcolor{gray!20}0.50 & 0.52 & \cellcolor{gray!20}0.50 & 0.75 & \cellcolor{gray!20}1.00 & 1.00 \\ 
163.1 & \cellcolor{gray!20}0.33 & 0.40 & \cellcolor{gray!20}0.44 & 0.57 & \cellcolor{gray!20}0.67 & 0.68 & \cellcolor{gray!20}0.33 & 0.51 & \cellcolor{gray!20}0.33 & 0.46 & \cellcolor{gray!20}1.00 & 1.00 & \cellcolor{gray!20}1.00 & 1.00 \\ 
231 & \cellcolor{gray!20}0.25 & 0.29 & \cellcolor{gray!20}0.38 & 0.54 & \cellcolor{gray!20}0.62 & 0.65 & \cellcolor{gray!20}0.44 & 0.51 & \cellcolor{gray!20}0.56 & 0.59 & \cellcolor{gray!20}0.94 & 0.94 & \cellcolor{gray!20}1.00 & 1.00 \\ 
249 & \cellcolor{gray!20}0.40 & 0.45 & \cellcolor{gray!20}0.33 & 0.41 & \cellcolor{gray!20}0.60 & 0.68 & \cellcolor{gray!20}0.66 & 0.74 & \cellcolor{gray!20}0.53 & 0.66 & \cellcolor{gray!20}0.87 & 0.91 & \cellcolor{gray!20}0.88 & 0.90 \\ 
254 & \cellcolor{gray!20}0.50 & 0.61 & \cellcolor{gray!20}0.65 & 0.73 & \cellcolor{gray!20}0.50 & 0.58 & \cellcolor{gray!20}0.40 & 0.52 & \cellcolor{gray!20}0.50 & 0.59 & \cellcolor{gray!20}0.75 & 0.85 & \cellcolor{gray!20}0.85 & 0.92 \\ 
264 & \cellcolor{gray!20}0.67 & 0.67 & \cellcolor{gray!20}0.50 & 0.56 & \cellcolor{gray!20}0.50 & 0.59 & \cellcolor{gray!20}0.42 & 0.51 & \cellcolor{gray!20}0.25 & 0.38 & \cellcolor{gray!20}0.92 & 0.96 & \cellcolor{gray!20}1.00 & 1.00 \\ 
267.12 & \cellcolor{gray!20}0.33 & 0.53 & \cellcolor{gray!20}0.33 & 0.44 & \cellcolor{gray!20}0.67 & 0.77 & \cellcolor{gray!20}0.75 & 0.84 & \cellcolor{gray!20}0.75 & 0.80 & \cellcolor{gray!20}1.00 & 1.00 & \cellcolor{gray!20}1.00 & 1.00 \\ 
267.5 & \cellcolor{gray!20}0.67 & 0.78 & \cellcolor{gray!20}0.75 & 0.85 & \cellcolor{gray!20}0.83 & 0.90 & \cellcolor{gray!20}0.67 & 0.76 & \cellcolor{gray!20}0.50 & 0.62 & \cellcolor{gray!20}1.00 & 1.00 & \cellcolor{gray!20}1.00 & 1.00 \\ 
267.8 & \cellcolor{gray!20}0.47 & 0.54 & \cellcolor{gray!20}0.56 & 0.62 & \cellcolor{gray!20}0.66 & 0.69 & \cellcolor{gray!20}0.56 & 0.63 & \cellcolor{gray!20}0.60 & 0.66 & \cellcolor{gray!20}0.83 & 0.87 & \cellcolor{gray!20}0.83 & 0.88 \\ 
268  & \cellcolor{gray!20}0.45 & 0.54 & \cellcolor{gray!20}0.25 & 0.41 & \cellcolor{gray!20}0.35 & 0.44 & \cellcolor{gray!20}0.35 & 0.44 & \cellcolor{gray!20}0.40 & 0.49 & \cellcolor{gray!20}0.50 & 0.65 & \cellcolor{gray!20}0.75 & 0.86 \\ 
279 & \cellcolor{gray!20}0.83 & 0.86 & \cellcolor{gray!20}0.92 & 0.92 & \cellcolor{gray!20}0.75 & 0.81 & \cellcolor{gray!20}0.83 & 0.88 & \cellcolor{gray!20}0.83 & 0.86 & \cellcolor{gray!20}1.00 & 1.00 & \cellcolor{gray!20}0.92 & 0.96 \\ 
380 & \cellcolor{gray!20}0.24 & 0.35 & \cellcolor{gray!20}0.24 & 0.36 & \cellcolor{gray!20}0.39 & 0.47 & \cellcolor{gray!20}0.47 & 0.53 & \cellcolor{gray!20}0.35 & 0.48 & \cellcolor{gray!20}0.78 & 0.80 & \cellcolor{gray!20}0.71 & 0.73 \\ 
462.37 & \cellcolor{gray!20}0.40 & 0.49 & \cellcolor{gray!20}0.40 & 0.52 & \cellcolor{gray!20}0.65 & 0.69 & \cellcolor{gray!20}0.67 & 0.70 & \cellcolor{gray!20}0.65 & 0.69 & \cellcolor{gray!20}0.78 & 0.80 & \cellcolor{gray!20}0.81 & 0.87 \\ 
465 & \cellcolor{gray!20}0.50 & 0.63 & \cellcolor{gray!20}0.75 & 0.76 & \cellcolor{gray!20}0.50 & 0.63 & \cellcolor{gray!20}0.38 & 0.54 & \cellcolor{gray!20}0.75 & 0.75 & \cellcolor{gray!20}1.00 & 1.00 & \cellcolor{gray!20}1.00 & 1.00 \\ 
467.1 & \cellcolor{gray!20}0.29 & 0.41 & \cellcolor{gray!20}0.57 & 0.75 & \cellcolor{gray!20}0.67 & 0.76 & \cellcolor{gray!20}0.33 & 0.64 & \cellcolor{gray!20}0.58 & 0.70 & \cellcolor{gray!20}1.00 & 1.00 & \cellcolor{gray!20}1.00 & 1.00 \\ 
495 & \cellcolor{gray!20}0.32 & 0.40 & \cellcolor{gray!20}0.32 & 0.46 & \cellcolor{gray!20}0.60 & 0.66 & \cellcolor{gray!20}0.56 & 0.61 & \cellcolor{gray!20}0.60 & 0.65 & \cellcolor{gray!20}0.77 & 0.87 & \cellcolor{gray!20}0.87 & 0.92 \\ 
530 & \cellcolor{gray!20}0.00 & 0.01 & \cellcolor{gray!20}0.00 & 0.01 & \cellcolor{gray!20}0.00 & 0.01 & \cellcolor{gray!20}0.00 & 0.02 & \cellcolor{gray!20}0.00 & 0.03 & \cellcolor{gray!20}1.00 & 1.00 & \cellcolor{gray!20}1.00 & 1.00 \\ 
591 & \cellcolor{gray!20}0.13 & 0.31 & \cellcolor{gray!20}0.25 & 0.36 & \cellcolor{gray!20}0.61 & 0.71 & \cellcolor{gray!20}0.52 & 0.58 & \cellcolor{gray!20}0.51 & 0.60 & \cellcolor{gray!20}0.87 & 0.93 & \cellcolor{gray!20}0.87 & 0.92 \\ 
601 & \cellcolor{gray!20}0.58 & 0.62 & \cellcolor{gray!20}0.58 & 0.64 & \cellcolor{gray!20}0.86 & 0.89 & \cellcolor{gray!20}0.79 & 0.81 & \cellcolor{gray!20}0.29 & 0.49 & \cellcolor{gray!20}0.86 & 0.93 & \cellcolor{gray!20}0.86 & 0.93 \\ 
650 & \cellcolor{gray!20}0.64 & 0.70 & \cellcolor{gray!20}0.72 & 0.77 & \cellcolor{gray!20}0.78 & 0.80 & \cellcolor{gray!20}0.69 & 0.74 & \cellcolor{gray!20}0.75 & 0.76 & \cellcolor{gray!20}0.97 & 0.99 & \cellcolor{gray!20}0.97 & 0.99 \\ 
672.73 & \cellcolor{gray!20}0.25 & 0.30 & \cellcolor{gray!20}0.25 & 0.32 & \cellcolor{gray!20}0.33 & 0.34 & \cellcolor{gray!20}0.33 & 0.34 & \cellcolor{gray!20}0.33 & 0.38 & \cellcolor{gray!20}0.67 & 0.71 & \cellcolor{gray!20}1.00 & 1.00 \\ 
672.78 & \cellcolor{gray!20}0.27 & 0.34 & \cellcolor{gray!20}0.34 & 0.43 & \cellcolor{gray!20}0.42 & 0.46 & \cellcolor{gray!20}0.50 & 0.55 & \cellcolor{gray!20}0.42 & 0.48 & \cellcolor{gray!20}0.83 & 0.92 & \cellcolor{gray!20}1.00 & 1.00 \\ 
676 & \cellcolor{gray!20}0.14 & 0.29 & \cellcolor{gray!20}0.14 & 0.27 & \cellcolor{gray!20}0.50 & 0.62 & \cellcolor{gray!20}0.57 & 0.66 & \cellcolor{gray!20}0.36 & 0.55 & \cellcolor{gray!20}0.93 & 0.94 & \cellcolor{gray!20}1.00 & 1.00 \\ 
683 & \cellcolor{gray!20}0.11 & 0.26 & \cellcolor{gray!20}0.18 & 0.30 & \cellcolor{gray!20}0.48 & 0.52 & \cellcolor{gray!20}0.52 & 0.57 & \cellcolor{gray!20}0.48 & 0.54 & \cellcolor{gray!20}0.81 & 0.88 & \cellcolor{gray!20}0.90 & 0.94 \\ 
684 & \cellcolor{gray!20}0.35 & 0.43 & \cellcolor{gray!20}0.60 & 0.72 & \cellcolor{gray!20}0.25 & 0.51 & \cellcolor{gray!20}0.25 & 0.34 & \cellcolor{gray!20}0.25 & 0.27 & \cellcolor{gray!20}1.00 & 1.00 & \cellcolor{gray!20}1.00 & 1.00 \\ 
686 & \cellcolor{gray!20}0.21 & 0.28 & \cellcolor{gray!20}0.28 & 0.36 & \cellcolor{gray!20}0.57 & 0.65 & \cellcolor{gray!20}0.65 & 0.68 & \cellcolor{gray!20}0.43 & 0.55 & \cellcolor{gray!20}0.68 & 0.79 & \cellcolor{gray!20}0.94 & 0.96 \\ 
687 & \cellcolor{gray!20}0.20 & 0.30 & \cellcolor{gray!20}0.30 & 0.49 & \cellcolor{gray!20}0.62 & 0.64 & \cellcolor{gray!20}0.38 & 0.51 & \cellcolor{gray!20}0.50 & 0.53 & \cellcolor{gray!20}0.88 & 0.94 & \cellcolor{gray!20}0.75 & 0.83 \\ 
715.1 & \cellcolor{gray!20}0.12 & 0.25 & \cellcolor{gray!20}0.12 & 0.22 & \cellcolor{gray!20}0.33 & 0.50 & \cellcolor{gray!20}0.33 & 0.45 & \cellcolor{gray!20}0.50 & 0.56 & \cellcolor{gray!20}1.00 & 1.00 & \cellcolor{gray!20}1.00 & 1.00 \\ 
718.1 & \cellcolor{gray!20}0.17 & 0.26 & \cellcolor{gray!20}0.08 & 0.24 & \cellcolor{gray!20}0.67 & 0.67 & \cellcolor{gray!20}0.67 & 0.67 & \cellcolor{gray!20}0.33 & 0.48 & \cellcolor{gray!20}1.00 & 1.00 & \cellcolor{gray!20}1.00 & 1.00 \\ 
718.2 & \cellcolor{gray!20}0.20 & 0.30 & \cellcolor{gray!20}0.17 & 0.31 & \cellcolor{gray!20}0.52 & 0.59 & \cellcolor{gray!20}0.52 & 0.57 & \cellcolor{gray!20}0.59 & 0.64 & \cellcolor{gray!20}0.76 & 0.85 & \cellcolor{gray!20}0.87 & 0.92 \\ 
784 & \cellcolor{gray!20}0.20 & 0.29 & \cellcolor{gray!20}0.30 & 0.49 & \cellcolor{gray!20}0.50 & 0.52 & \cellcolor{gray!20}0.38 & 0.46 & \cellcolor{gray!20}0.50 & 0.52 & \cellcolor{gray!20}1.00 & 1.00 & \cellcolor{gray!20}1.00 & 1.00 \\ 
839 & \cellcolor{gray!20}0.17 & 0.25 & \cellcolor{gray!20}0.07 & 0.20 & \cellcolor{gray!20}0.33 & 0.37 & \cellcolor{gray!20}0.33 & 0.36 & \cellcolor{gray!20}0.33 & 0.35 & \cellcolor{gray!20}1.00 & 1.00 & \cellcolor{gray!20}1.00 & 1.00 \\  

\cmidrule(r){2-15}

\textbf{Average} & \cellcolor{gray!20}0.36 & 0.45 & \cellcolor{gray!20}0.40 & 0.50 & \cellcolor{gray!20}0.53 & 0.59 & \cellcolor{gray!20}0.45 & 0.54 & \cellcolor{gray!20}0.46 & 0.53 & \cellcolor{gray!20}0.77 & 0.83 & \cellcolor{gray!20}0.86 & 0.90 \\

    \end{tabular}
    }
    \caption{Results on the `Criminal Code Sections (Canada)' \textsc{LegalLAMA} task. We kept only the sections with more than one example.}
    \label{tab:lama_ccc_term}
\end{table*}

\section{LegalLAMA Tasks' Vocabulary}
\label{sec:legalama_lists}

In Tables~\ref{tab:lama_echr}, \ref{tab:lama_contract_sections}, \ref{tab:lama_contract_types}, \ref{tab:lama_eu_term}, and~\ref{tab:lama_ccc_term} we present the labels' list for the `ECHR Articles', `Contract Sections', `Contract Types', `Terminology (EU)' and 'Criminal Code Sections (Canada)' sub-tasks and the label-wise performance. In Tables~\ref{tab:coe_terms}, \ref{tab:crime_charges_terms}, and \ref{tab:us_terms},  we present the labels' list for the `Terminology (CoE)', `Crimes Charges (US)', and `Terminology (US)' sub-tasks grouped in clusters.

\begin{table*}
\centering
    \resizebox{\textwidth}{!}{
    \small
    \begin{tabular}{l|p{14cm}}
    \toprule
    \bf ECHR Article & \bf Masked Terms \\
    \midrule
Art. 2 & `accessibility' , `effective investigation' , `expulsion' , `extradition' , `foreseeability' , `positive obligations' , `prescribed by law' , `right to life' , `safeguards against abuse' , `use of force' \\
\midrule
Art. 3 & `effective investigation' , `expulsion' , `extradition' , `inhuman punishment' , `inhuman treatment' , `positive obligations' , `prohibition of torture' , `torture' \\
\midrule
Art. 5 & `competent court' , `deprivation of liberty' , `drug addicts' , `educational supervision' , `expulsion' , `extradition' , `guarantees to appear for trial' , `lawful arrest or detention' , `lawful order of a court' , `length of pre-trial detention' , `minors' , `order release' , `persons of unsound mind' , `procedure prescribed by law' , `reasonable suspicion' , `release pending trial' , `review by a court' , `right to liberty and security' , `security of person' , `speediness of review' , `take proceedings' , `trial within a reasonable time' \\
\midrule
Art. 6 & `charged with a criminal offence' , `disciplinary proceedings' , `enforcement proceedings' , `equality of arms' , `examination of witnesses' , `exclusion of public' , `expulsion' , `extradition' , `fair hearing' , `free legal assistance' , `impartial tribunal' , `independent tribunal' , `insufficient means' , `legal aid' , `national security' , `necessary in a democratic society' , `oral hearing' , `presumption of innocence' , `protection of public order' , `proved guilty according to law' , `public hearing' , `public judgment' , `reasonable time' , `right to a fair trial' , `rights of defence' , `same conditions' , `tribunal established by law' \\
\midrule
Art. 7 & `criminal offence' , `heavier penalty' , `retroactivity' \\
\midrule
Art. 8 & `accessibility' , `economic well-being of the country' , `expulsion' , `extradition' , `foreseeability' , `interference' , `national security' , `necessary in a democratic society' , `positive obligations' , `prevention of crime' , `prevention of disorder' , `protection of health' , `protection of morals' , `protection of the rights and freedoms of others' , `public authority' , `public safety' , `respect for correspondence' , `respect for family life' , `respect for home' , `respect for private life' , `right to respect for private and family life' , `safeguards against abuse' \\
\midrule
Art. 9 & `foreseeability' , `freedom of conscience' , `freedom of religion' , `freedom of thought' , `interference' , `necessary in a democratic society' , `observance' , `positive obligations' , `practice' , `prescribed by law' , `protection of health' , `protection of public order' , `protection of the rights and freedoms of others' , `public safety' , `safeguards against abuse' , `teaching' , `worship' \\
\midrule
Art. 10 & `duties and responsibilities' , `foreseeability' , `freedom of expression' , `freedom to hold opinions' , `freedom to impart information' , `freedom to receive information' , `interference' , `national security' , `necessary in a democratic society' , `positive obligations' , `prescribed by law' , `prevention of crime' , `prevention of disorder' , `protection of health' , `protection of morals' , `protection of the reputation of others' , `protection of the rights of others' , `public safety' , `safeguards against abuse' , `territorial integrity' \\
\midrule
Art. 11 & `accessibility' , `foreseeability' , `form and join trade unions' , `freedom of assembly and association' , `freedom of association' , `freedom of peaceful assembly' , `interference' , `national security' , `necessary in a democratic society' , `positive obligations' , `prescribed by law' , `prevention of crime' , `prevention of disorder' , `protection of health' , `public safety' \\
\midrule
Art. 13 & `effective remedy' , `national authority' , `right to an effective remedy' \\
\midrule
Art. 14 & `discrimination' , `language' , `national minority' , `national origin' , `objective and reasonable justification' , `prohibition of discrimination' , `property' , `race' , `religion' , `sex' , `social origin' \\
\midrule
Art. 35 & `continuing situation' , `effective domestic remedy' , `exhaustion of domestic remedies' , `final domestic decision' , `manifestly ill-founded' , `no significant disadvantage' , `relevant new information' \\
\midrule
Art. P1-1 & `accessibility' , `deprivation of property' , `foreseeability' , `general interest' , `general principles of international law' , `interference' , `peaceful enjoyment of possessions' , `positive obligations' , `possessions' , `prescribed by law' , `protection of property' , `secure the payment of taxes' \\
\bottomrule
\end{tabular}
}
\caption{Masked Terms used in the `Terminology (CoE)' \textsc{LegalLAMA} task.}
\label{tab:coe_terms}
\end{table*}

\begin{table*}
\centering
    \resizebox{\textwidth}{!}{
    \small
    \begin{tabular}{l|p{14cm}}
    \toprule
    \bf Crime Area & \bf Masked Terms \\
    \midrule
Children & `child abandonment' , `child abuse' \\
\midrule
Computer & `computer crime' , `cyberbullying' , `identity theft' \\
\midrule
Court-related & `criminal contempt of court' , `perjury' , `probation violation' \\
\midrule
Drug-related & `drug distribution' , `drug manufacturing' , `drug possession' , `drug trafficking' , `medical marijuana' , `minor in possession' , `public intoxication' \\
\midrule
Life Taking & `homicide' , `manslaughter' , `murder' \\
\midrule
Mens Rea & `accessory' , `aiding and abetting' , `attempt' , `conspiracy' , `hate crime' \\
\midrule
Monetary & `bribery' , `embezzlement' , `extortion' , `forgery' , `insurance fraud' , `money laundering' , `pyramid schemes' , `racketeering' , `securities fraud' , `shoplifting' , `tax evasion' , `telemarketing fraud' , `theft' , `white collar crime' , `wire fraud' \\
\midrule
Behavior & `disorderly conduct' , `disturbing the peace' , `harassment' , `stalking' \\
\midrule
Property & `arson' , `vandalism' \\
\midrule
Sex-related & `child pornography' , `indecent exposure' , `prostitution' , `rape' , `sexual assault' , `solicitation' , `statutory rape' \\
\midrule
Violence & `aggravated assault' , `battery' , `burglary' , `domestic violence' , `kidnapping' , `robbery' \\
\bottomrule
\end{tabular}
}
\caption{Masked Terms used in the `Crime Charges (US)' \textsc{LegalLAMA} task grouped by crime areas.}
\label{tab:crime_charges_terms}
\end{table*}

\begin{table*}
\centering
    \resizebox{\textwidth}{!}{
    \small
    \begin{tabular}{p{2cm}|p{14cm}}
    \toprule
    \bf Legal Topic & \bf Masked Terms \\
    \midrule
Business Law & `adhesion contract' , `implied warranty' , `limited liability' , `parol evidence' , `quantum meruit' , `reliance damages' , `self-dealing' , `severability clause' , `specific performance' , `statute of frauds' , `substantial performance' , `tender offer' , `third-party beneficiary' , `unconscionability' \\
\midrule
Criminal Law and Procedure & `accessory before the fact' , `accomplice' , `aggravated assault' , `allocution' , `arson' , `defense of others' , `inchoate' , `merger doctrine' , `mitigating circumstances' , `money laundering' , `stop and frisk' \\
\midrule
Employment Law & `bargaining unit' , `boycott' , `casual labor' , `industrial safety' , `minimum wage' , `workplace safety' , `wrongful termination' \\
\midrule
Family Law & `consent divorce' , `emancipation of minors' , `marital privilege' , `marital property' , `marital settlement agreement' , `separate property' , `separation agreement' , `shared custody' , `sole custody' , `spousal privilege' , `spousal support' , `visitation' , `wage attachment' \\
\midrule
Immigration & `alienage' , `asylum seeker' , `asylum' , `childhood arrivals' , `citizenship' , `deferred action' , `deportation' , `geneva conventions' , `naturalization' , `nonresident' , `refugee' , `resettlement' , `visa' \\
\midrule
Landlord-Tenant Law & `abandonment' , `commercial reasonability' , `constructive eviction' , `eviction' , `habitability' , `privity' , `quiet enjoyment' , `reasonableness' , `self-help eviction' , `sole discretion' , `tenancy at sufferance' , `tenancy at will' \\
\midrule
Money And Financial Problems & `bankruptcy discharge' , `bond' , `consumer credit' , `kiting' , `malfeasance' , `mortgage' , `nonrecourse' , `ponzi scheme' , `securities fraud' , `self-dealing' , `senior lien' , `stock dividend' , `straw man' , `swindle' , `tontine' , `variable annuity' \\
\bottomrule
\end{tabular}
}
\caption{Masked Terms used in the `Terminology (US)' \textsc{LegalLAMA} task grouped by legal topics.}
\label{tab:us_terms}
\end{table*}

\end{document}